\documentclass[conference]{IEEEtran}

\usepackage{amsmath,amsfonts}
\usepackage{array}

\usepackage{subcaption}

\usepackage{stackengine}

\usepackage{textcomp} 
\usepackage{stfloats}
\usepackage[hyphens]{url}
\usepackage{verbatim}
\usepackage{graphicx}
\usepackage[dvipsnames]{xcolor}
\usepackage[numbers]{natbib}
\usepackage{times}
\usepackage{multicol}
\PassOptionsToPackage{hyphens}{url}\usepackage{hyperref}
\usepackage[normalem]{ulem}
\usepackage{arydshln}

\let\labelindent\relax
\usepackage{enumitem}
\usepackage{tabularx}

\usepackage{booktabs}       

\usepackage{arydshln}
\makeatletter
\def\adl@drawiv#1#2#3{%
        \hskip.5\tabcolsep
        \xleaders#3{#2.5\@tempdimb #1{1}#2.5\@tempdimb}%
                #2\z@ plus1fil minus1fil\relax
        \hskip.5\tabcolsep}
\newcommand{\cdashlinelr}[1]{%
  \noalign{\vskip\aboverulesep
           \global\let\@dashdrawstore\adl@draw
           \global\let\adl@draw\adl@drawiv}
  \cdashline{#1}
  \noalign{\global\let\adl@draw\@dashdrawstore
           \vskip\belowrulesep}}
\makeatother

\usepackage{algorithm}
\usepackage[noend]{algpseudocode}
\definecolor{commentcolor}{gray}{0.5}
\algnewcommand{\LineComment}[1]{\State \textcolor{commentcolor}{\(\triangleright\) #1}}
\algnewcommand{\var}[1]{\textit{#1}}
\algnewcommand{\func}[1]{\textsc{#1}}

\def\papertitle{
Demonstrating Multi-Suction Item Picking at Scale via Multi-Modal Learning of Pick Success}

\begin{document}

\title{
\papertitle
}

\author{\authorblockN{Che Wang\authorrefmark{1}, Jeroen van Baar\authorrefmark{2}, Chaitanya Mitash\authorrefmark{2}, Shuai Li\authorrefmark{1}, Dylan Randle\authorrefmark{2}, \\ Weiyao Wang\authorrefmark{2}, Sumedh Sontakke\authorrefmark{1},  Kostas E. Bekris\authorrefmark{2}, and Kapil Katyal\authorrefmark{3}}
\authorblockA{Amazon Robotics\\\authorrefmark{1}Seattle, WA; \authorrefmark{2}North Reading, MA; \authorrefmark{3}Arlington, VA\\Email: (corresponding author) chewwang@amazon.com}}

\twocolumn[{
\begin{@twocolumnfalse}

\maketitle
\captionsetup{type=figure}\includegraphics[width=\linewidth]{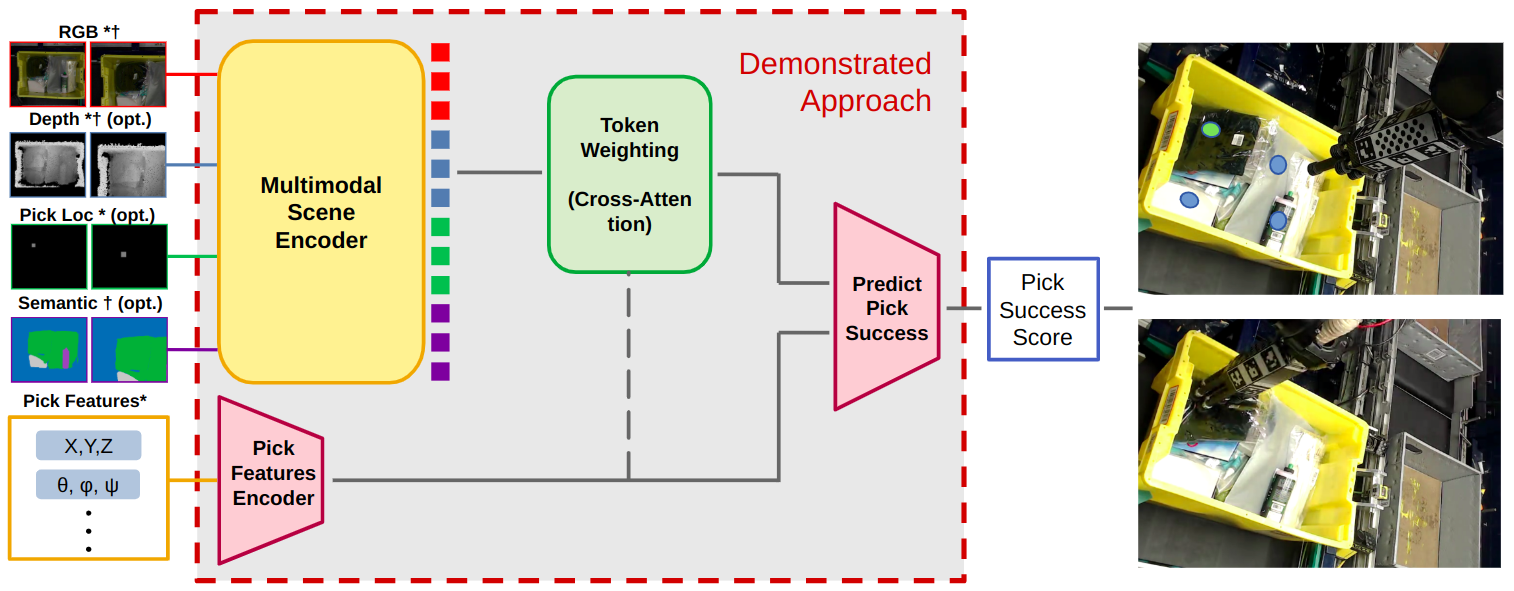}
\captionof{figure}{Given RGB, depth, pick location and semantic data for a picking scene as well as features of candidate, multi-suction picks, this work demonstrates how state-of-the-art multi-modal visual encoders can learn expressive representations. These representations are used to evaluate candidate picks via a cross-attention mechanism and a pick success prediction head. All input image modalities on the left (RGB, Depth, Pick Location, Semantic) can also be cropped to produce a local image (right side of the input images pairs). Operating over local image crops helps boost performance. The modalities marked with $^*$ are used in the default approach. The modalities marked with $^\dagger$ are utilized during pretraining. The modalities marked with (opt.) are optional during the finetuning/inference but can enable higher performance if deployed. The demonstrated strategy is trained using picks executed on an operating, industrial setup. They are generated by a previously deployed engineered approach. The strategy achieves improved performance relative to the engineered approach and learning-based alternatives.  \vspace{0.1in}}
\label{fig:model-schematic}
\end{@twocolumnfalse}
}]

\begin{abstract}
This work demonstrates how autonomously learning aspects of robotic operation from sparsely-labeled, real-world data of deployed, engineered solutions at industrial scale can provide with solutions that achieve improved performance.  Specifically, it focuses on multi-suction robot picking and performs a comprehensive study on the application of multi-modal visual encoders for predicting the success of candidate robotic picks.  Picking diverse items from unstructured piles is an important and challenging task for robot manipulation in real-world settings, such as warehouses. Methods for picking from clutter must work for an open set of items while simultaneously meeting latency constraints to achieve high throughput. The demonstrated approach utilizes multiple input modalities, such as RGB, depth and semantic segmentation, to estimate the quality of candidate multi-suction picks. The strategy is trained from real-world experience, i.e., given examples of successful and failed attempts to pick items. The training picks have been generated by an engineered strategy. A real-world limitation when learning in such live, industrial setups is that only a single or a few picks can be attempted per scene. The learning strategy first pretrains multi-modal visual models in a self-supervised manner to effectively reconstruct the input modalities in the target domain. A downstream model is then trained to evaluate the quality of multi-suction picks given the learned multi-modal embedding, while the multi-modal model is further fine-tuned. The manuscript provides comprehensive experimental evaluation performed over a large item-picking dataset, an item-picking dataset targeted to include partial occlusions, and a package-picking dataset, which focuses on containers, such as boxes and envelopes, instead of unpackaged items. The evaluation measures performance for different item configurations, pick scenes, and object types. Ablations help to understand the effects of in-domain pretraining, the impact of different modalities and the importance of finetuning. These ablations reveal both the importance of training over multiple modalities but also the ability of models to learn during pretraining the relationship between modalities so that during finetuning and inference, only a subset of them can be used as input.

\end{abstract}

\IEEEpeerreviewmaketitle

\section{Introduction} \label{sec:intro}

Picking items from unstructured piles is an important yet challenging task for robotic manipulation in real-world settings, such as warehouse automation. Methods for picking from clutter have to be robust to an open set of items while simultaneously meeting latency constraints to achieve high throughput. While robot manipulation in clutter has long been approached via model-based reasoning \cite{bayes_grasp, contact_reactive, dogar_sri_planning, sukhatme_motion_primitives}, the recent focus has been on data-driven approaches and their potential benefits. In particular, learning-based methods have been introduced to address the large variety of items, for pinch grasping~\cite{googlepick, mahler2017dex, fang2023anygrasp}, suction-based grasping~\cite{mahler2017binpicking, suctionnet}, or both~\cite{, mahler2018dex, mahler2019learning}. Suction-based grasping, the modality considered in this work, is popular in real-world settings (e.g., logistics and fulfillment) as it simplifies the attachment of a target object to a robot's end-effector.

This work is inspired by data-driven methods for robot picking as well as the progress in multimodal, multi-task models~\cite{yuan2023m2t2}. It aims to explore the impact of such models in robot picking given the availability of large-scale data obtained from real-world industrial robotic deployments. 
The accompanying experiments use large real-world picking datasets.
In contrast to prior work that is often limited either to a closed set of items or training in simulation, this work aims to address the case of picking an open-set of possible items that can appear in any possible configuration, i.e., all possible products and configurations that arise in a real-world warehouse environment, using a multi-suction end-effector. 

Recent work~\cite{Li:2023:rss:pickranking} proposed a shallow model for pick success evaluation, which relied on  features engineered by human experts. The model was similarly derived from large-scale, in-domain multimodal data for a multi-suction end-effector. It focused, however, on picking packages instead of items from an open-set. The variety in appearance for packages, such as boxes, or envelopes, is considerably smaller compared to an open-set of items, considered here.  The prior work demonstrated that the shallow model with engineered features achieved stronger performance than a deep learning model\footnote{That model used a convolutional visual encoder using as input an RGB image but without access to  multimodal data or engineered features.}.

The current study first confirms the prior finding in the context of open-set item picking, i.e., it is not trivial to develop a deep architecture that learns effective visual representations that outperform a shallow model using features engineered by an expert (see Fig.~\ref{fig:early}). Motivated, however, by the need to automate and simplify the development and training of picking solutions,
this work demonstrates that recent deep visual architectures allow the automated learning of multimodal representations from real-world data. 

In particular, we propose a model for pick success prediction (see Fig. \ref{fig:model-schematic}) that operates over a multimodal visual encoder, such as the Multimodal Multi-Task Masked Autoencoder (MultiMAE)~\cite{bachmann2022multimae} trained on RGB, depth and semantic data. This encoder can be pretrained to correlate information across different modalities, resulting in a representation that captures information about items and their relation in a cluttered pile. Then, the learned multi-modal representation of the scene is combined with pick features via a cross-attention mechanism and is finetuned to predict the quality of the pick. Experiments show that this architecture achieves improved performance over a highly-tuned, shallow model as well as against deep model baselines that utilize a generic, frozen encoder. The demonstrated architecture also consistently outperforms the alternatives when applied to different types of picks as well as to scenes corresponding to package picking. All experiments presented are focused on multi-suction end effectors as they work well for the warehouse setting. 

\noindent In summary, the contributions of this work are the following: 

1. The combination of multimodal pretraining and finetuning of the MultiMAE, properly coupled with an encoding of the pick candidate information, such as position and orientation, replaces the need for engineered features, and outperforms the previously demonstrated shallow model.

2. We present experiments showing the learned multimodal representation works on
a standard item picking setting, item picking with increased occlusion and random pick samples, and a setting where we pick packages instead of items. 

3. Extensive ablations of the demonstrated architecture reveal which technical components are critical to achieving the best performance and a series of insights in the large-scale robot picking setting.
For instance, while many efforts in the literature advocate for frozen visual representations trained on large generic datasets~\cite{nair2022r3m, parisi2022unsurprising, xiao2022masked}, we show that considerable performance gains can be achieved through in-domain pretraining and finetuning. The majority of these gains can be achieved even with small exposure to in-domain multimodal pretraining (e.g., just seeing 1\% data of the available data).

In the appendix, we provide additional results, in the supplementary materials, we provide videos of picking items from clutter to illustrate the challenging nature of the task.

\section{Related Work} \label{sec:related-work}

This section first reviews work on robot picking and then discusses learning visual representations for robotic tasks.

\subsection{Robotic Item Picking} \label{sec:related-work:methods}

Traditional methods for robot grasping \cite{graspit} involve geometric reasoning, planning and optimization methods. They often calculate the poses and forces for robotic contacts so as to satisfy certain mechanical constraints, such as force or form closure \cite{murray1994mathematical, mason2001mechanics}. The requirement, however, for accurate geometric and physical object models often limits the effectiveness of these solutions in unstructured setups involving an open-set of objects, which is the focus of this work.

This motivated the introduction of learning-based approaches for robot grasping more than a decade ago \cite{dl_robot_grasping}. Such data-driven methods brought the promise of more effective grasping in challenging, cluttered setups with unknown objects. The majority of these works focus on generating grasps for parallel, pinch grippers \cite{fang2023anygrasp, fang2020graspnet, yuan2023m2t2}. Since the Amazon Picking Challenge \cite{apc_paper}, however, it has been well understood that suction-based grippers can be very effective in real-world picking setups by simplifying pick reasoning.  While the underlying representation learning tools and insights of this work are not necessarily limited only to such end-effectors, the accompanying experiments have been performed using a multi-suction gripper. Suction-based grippers have been deployed in real-world production environments and allow fast and robust picking of items that can potentially be heavy. 

Some of the data-driven solutions for robot picking have been extended to address suction-based grippers, such as the Dex-Net family of solutions \cite{mahler2018dex, suctionnet}. In Dex-Net, a pick candidate is evaluated using an expert-designed evaluation system. In this work, we directly train the model to learn pick success prediction given success labels of past production data, and demonstrate that recent deep architectures have the ability to generate informative representations for this task. This direction minimizes the need for expert knowledge, giving a more effective solution to the picking problem.

In the context of approaches that have been tested on large-scale, real-world data, prior work \citet{Li:2023:rss:pickranking} has demonstrated that a shallow model can have good performance in terms of pick success prediction as part of a larger grasp planning pipeline for package picking. Packages tend to have more limited geometries and physical features relative to picking an open set of items, which is the focus of this work. The shallow model in prior work outperformed a deep learning model with a convolutional visual encoder. This work shows that with proper pretraining and finetuning, a deep model with a multimodal visual encoder can in fact outperform the shallow model as well as alternative deep architectures in terms of picking items from an open-set.  Thus, a critical objective is to identify appropriate intermediate visual representations that allow the evaluation of robotic picks.

\subsection{Learning Visual Representations} \label{sec:related-work:representations}

The value of a visual representation for control has been recognized previously. Pretrained vision models for control~\cite{parisi2022unsurprising} exploit a visual representation for learning a motor policy. Alternatives focused on manipulation~\cite{nair2022r3m} have proposed a representation for video image data for learning a variety manipulation tasks.  A Masked Auto-Encoder (MAE)~\cite{He2021MaskedAA}  was proposed to be used to learn a control policy~\cite{xiao2022masked}. In order to exploit the multimodal nature of the available data, this work adopts Multi-Modal Multi-Task Masked Autoencoder (MultiMAE)~\cite{bachmann2022multimae}, the multimodal extension of MAE instead. In addition, the above mentioned approaches use generic robotic data for pretraining and freeze the visual encoder during downstream operation. This work identifies that when large amounts of in-domain data are available, pretraining and finetuning in-domain data provides improved performance. 

There are also other recent efforts that investigate using multimodal input for robotic control \cite{Jaegle2021PerceiverIA, yuan2023m2t2, karamcheti2023language, reed2022generalist}. We focus on MultiMAE due to: 

(1) Multiple visual modalities are often available in robotic workcells: a color camera for RGB data, a depth sensor for depth images, and a segmentation model that provides semantic segmentation information. To generate the robot picking actions, the demonstrated architecture reasons about the 3D surface of the items as well as the segment boundaries of the items. So we expect additional modalities including depth and segment can help the model to better predict pick outcomes. The accompanying ablations show that multimodal pretraining and finetuning (with RGB, depth and semantics), which MultiMAE can leverage~\cite{bachmann2022multimae}, bring significant performance improvements compared to using a single modality. 

(2) We are interested in a representation that can be pretrained in a self-supervised manner. This allows us to operate over large-scale data without 
additional curation and labeling. MAEs lend themselves well to this desirable objective. On top of that, they are also simple, robust and efficient \cite{He2021MaskedAA}.

To the best of our knowledge, this is the first application of a multimodal model for prediction of pick success at scale.

\section{Setup} 

Robotic work cells in the industry are instrumented with sensors of various modalities, such as color and depth. The sensors are typically providing a top-down observation of the cluttered items to be picked. An example top-down view is shown in Fig.~\ref{fig:model-schematic}. The visible items in the color images are segmented using previously trained segmentation models, providing semantic information. A set of pick candidate locations is generated across the items, where a pick candidate is defined by a number of ``pick features'', including $X, Y, Z$ coordinates of the pick location, which pistons of the multi-suction gripper are activated during the pick, orientation of the end-effector, among others. To achieve high throughput, items with a high probability of pick success for a particular pick candidate, should be picked. The goal is thus to train a pick success prediction model, which given the multimodal image data, provides a prediction of success for a given pick candidate.

\begin{figure*}[htb]
\captionsetup[subfigure]{justification=centering}
\centering
\begin{subfigure}[t]{.24\linewidth}
\centering
\includegraphics[width=\linewidth]{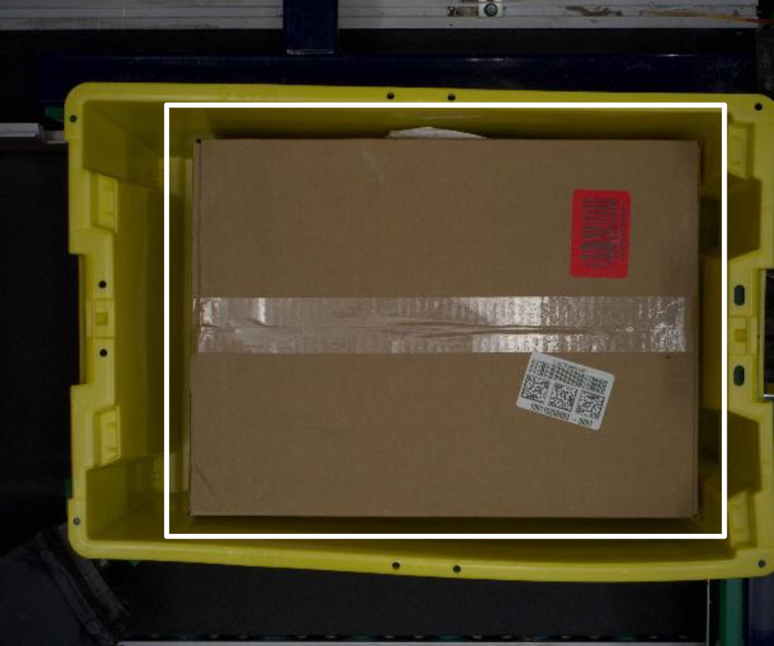}
\caption{Big heavy box}
\end{subfigure}
\begin{subfigure}[t]{.24\linewidth}
\centering
\includegraphics[width=\linewidth]{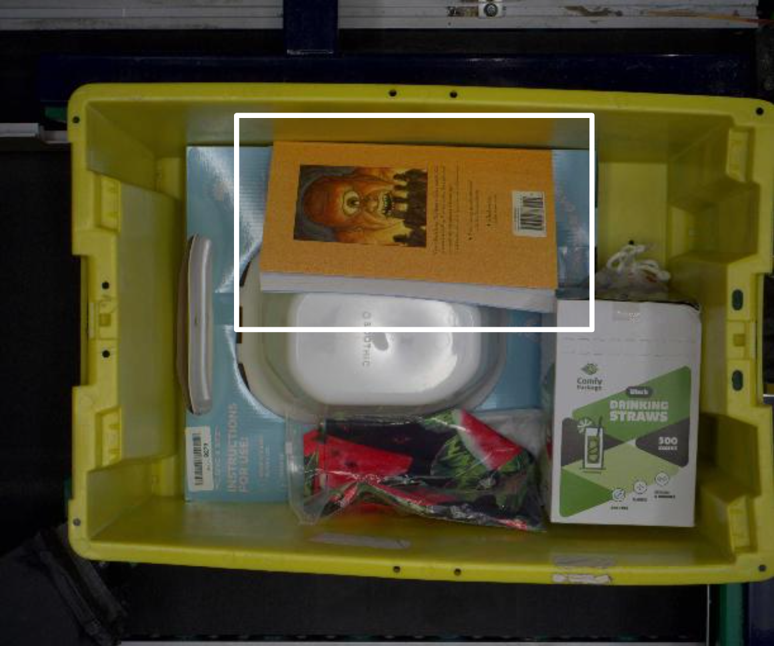}
\caption{Book; irregular shaped item}
\end{subfigure}
\begin{subfigure}[t]{.24\linewidth}
\centering
\includegraphics[width=\linewidth]{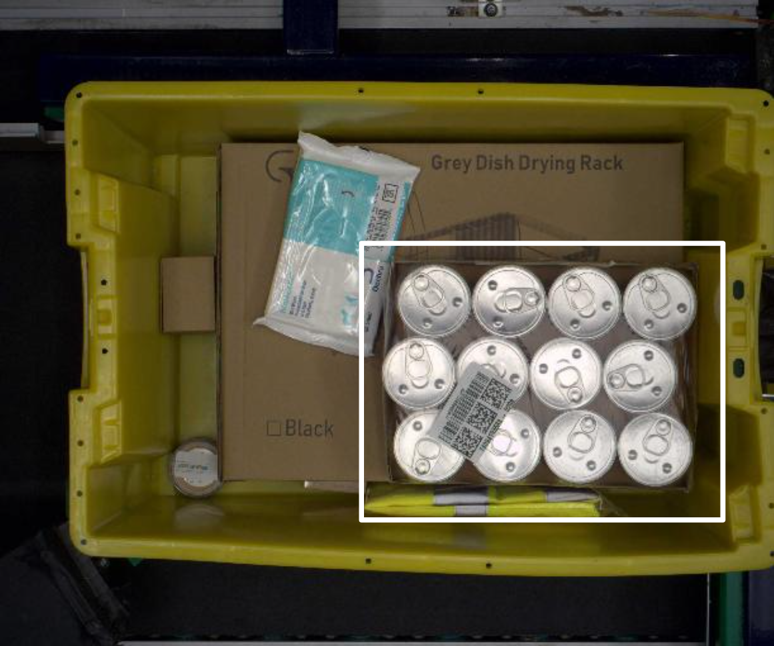}
\caption{Pack of cans}
\end{subfigure}
\begin{subfigure}[t]{.24\linewidth}
\centering
\includegraphics[width=\linewidth]{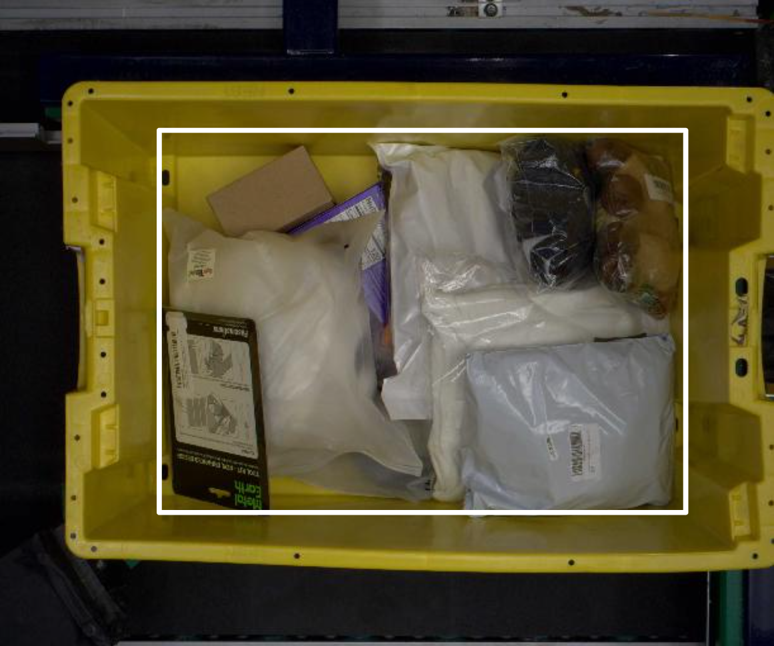}
\caption{Deformable bags}
\end{subfigure}\\

\begin{subfigure}[t]{.24\linewidth}
\centering
\includegraphics[width=\linewidth]{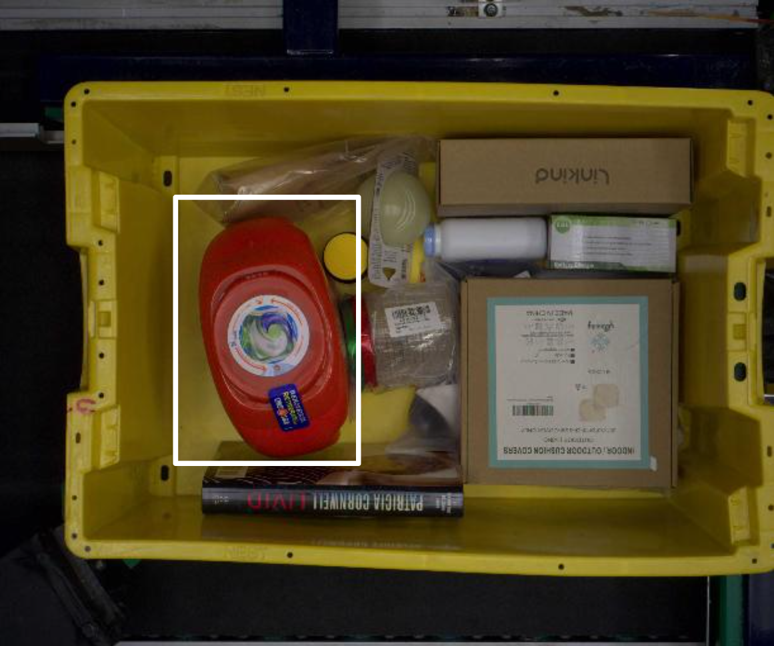}
\caption{Washer pods container}
\end{subfigure}
\begin{subfigure}[t]{.24\linewidth}
\centering
\includegraphics[width=\linewidth]{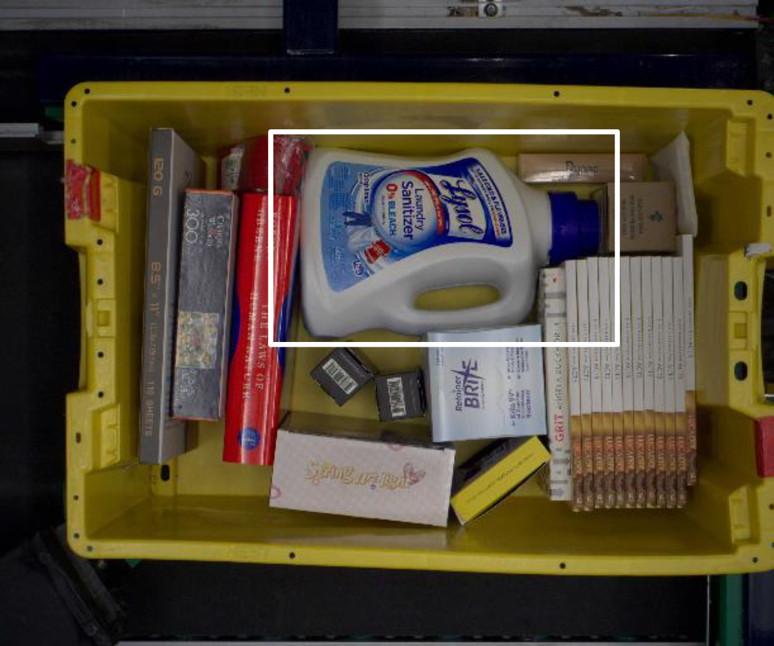}
\caption{Detergent}
\end{subfigure}
\begin{subfigure}[t]{.24\linewidth}
\centering
\includegraphics[width=\linewidth]{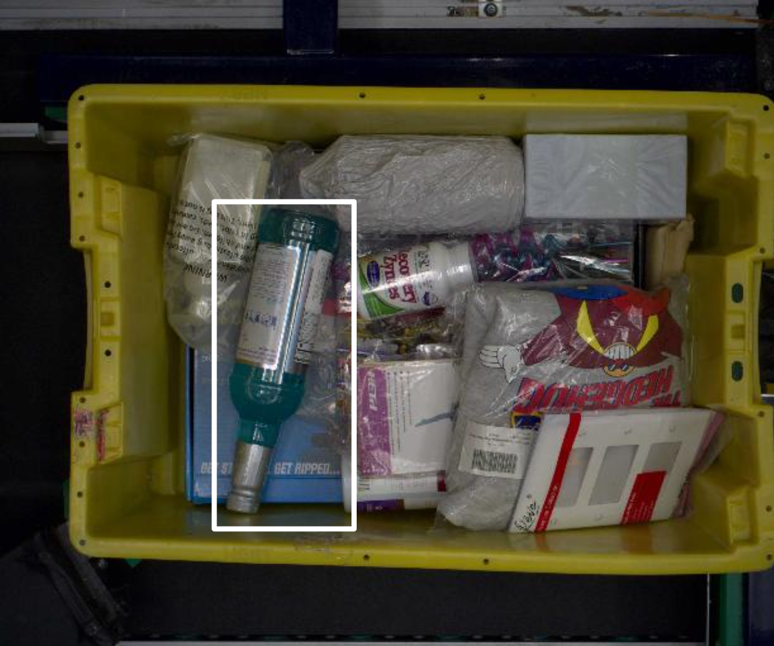}
\caption{Glass bottles}
\end{subfigure}
\begin{subfigure}[t]{.24\linewidth}
\centering
\includegraphics[width=\linewidth]{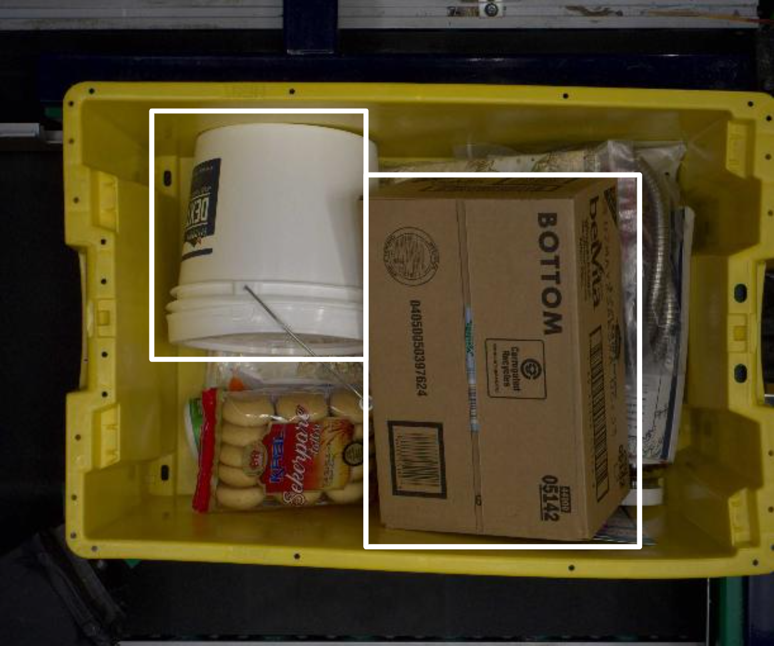}
\caption{Bucket; bag of pastry}
\end{subfigure}\\

\begin{subfigure}[t]{.24\linewidth}
\centering
\includegraphics[width=\linewidth]{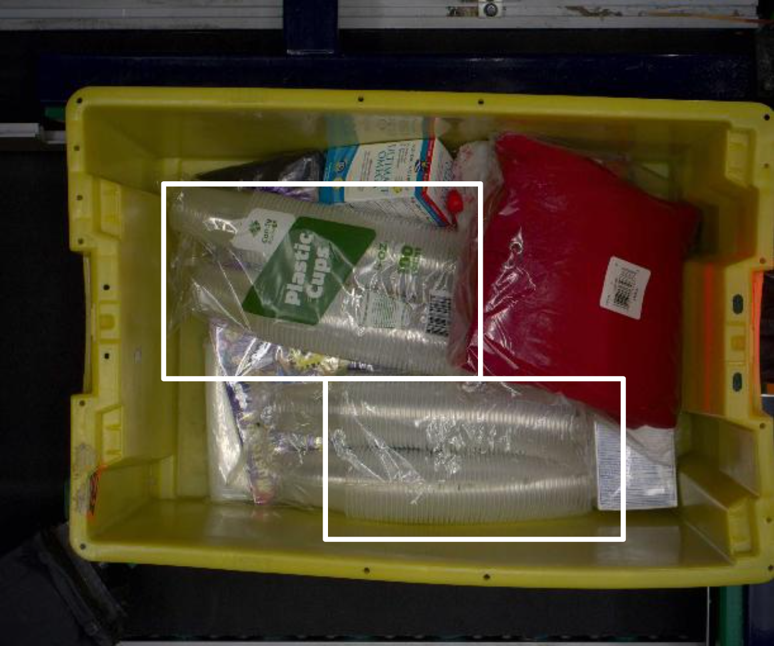}
\caption{Plastic cups}
\end{subfigure}
\begin{subfigure}[t]{.24\linewidth}
\centering
\includegraphics[width=\linewidth]{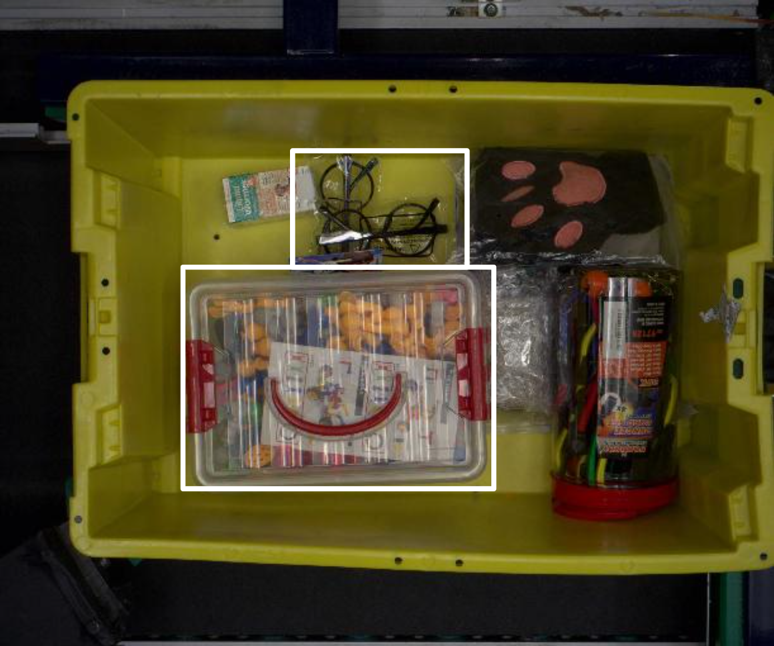}
\caption{Transparent box; glasses}
\end{subfigure}
\begin{subfigure}[t]{.24\linewidth}
\centering
\includegraphics[width=\linewidth]{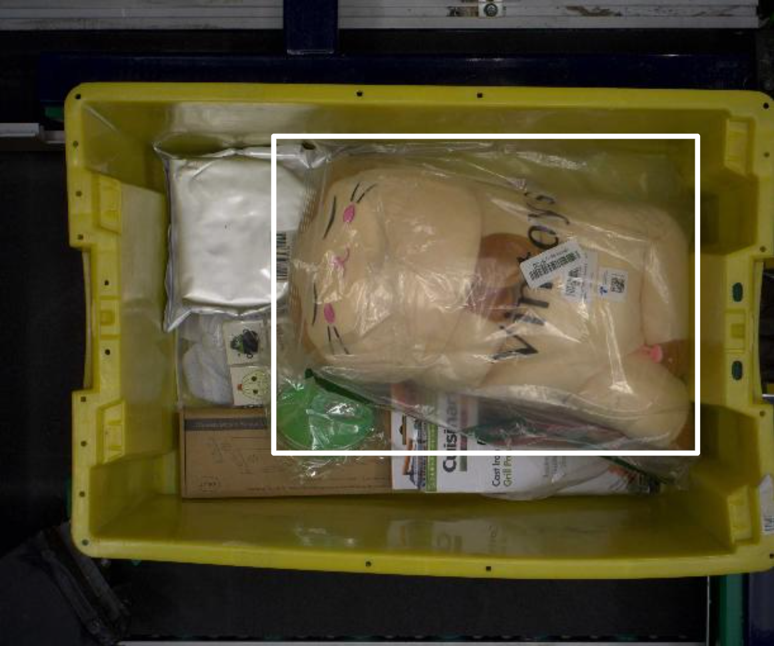}
\caption{Plush}
\end{subfigure}
\begin{subfigure}[t]{.24\linewidth}
\centering
\includegraphics[width=\linewidth]{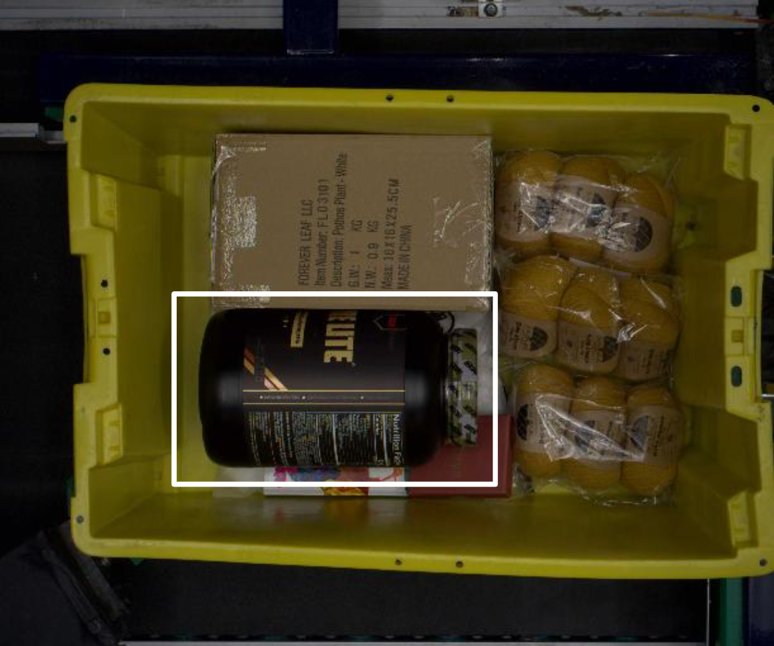}
\caption{Large protein powder bottle}
\end{subfigure}\\

\begin{subfigure}[t]{.24\linewidth}
\centering
\includegraphics[width=\linewidth]{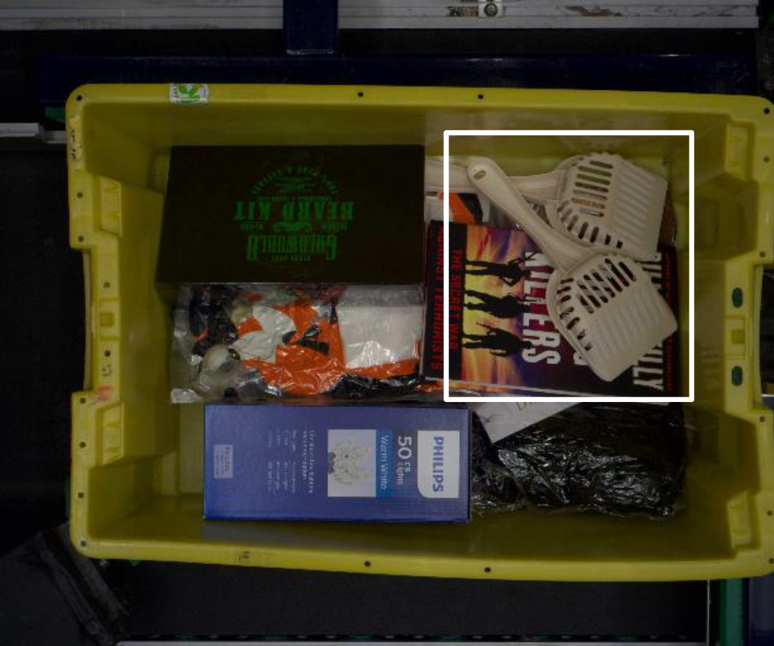}
\caption{Spatula without packaging}
\end{subfigure}
\begin{subfigure}[t]{.24\linewidth}
\centering
\includegraphics[width=\linewidth]{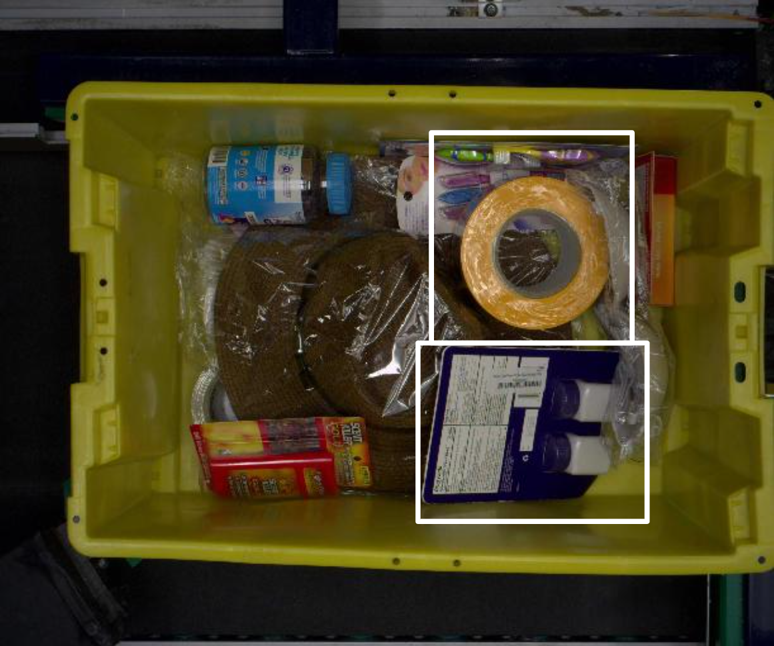}
\caption{Tape; irregular shaped item}
\end{subfigure}
\begin{subfigure}[t]{.24\linewidth}
\centering
\includegraphics[width=\linewidth]{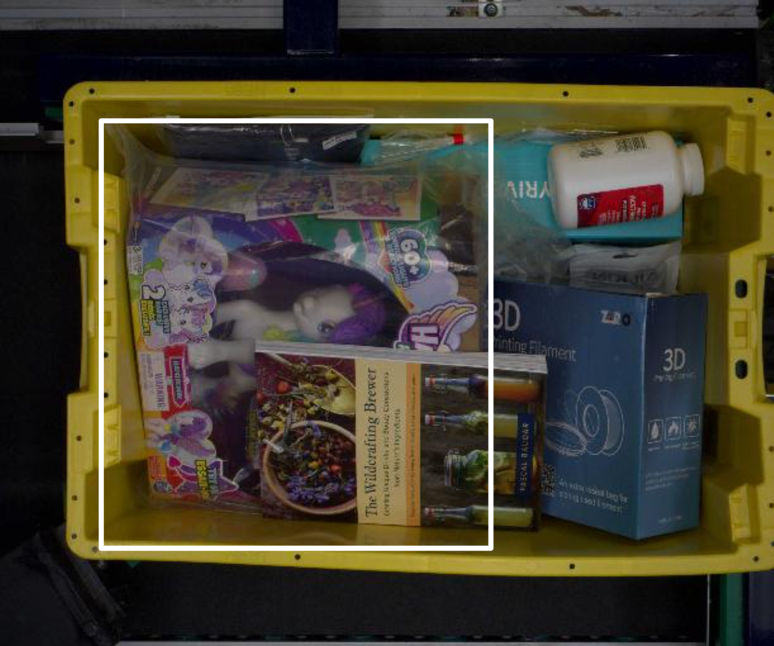}
\caption{Transparent toy box}
\end{subfigure}
\begin{subfigure}[t]{.24\linewidth}
\centering
\includegraphics[width=\linewidth]{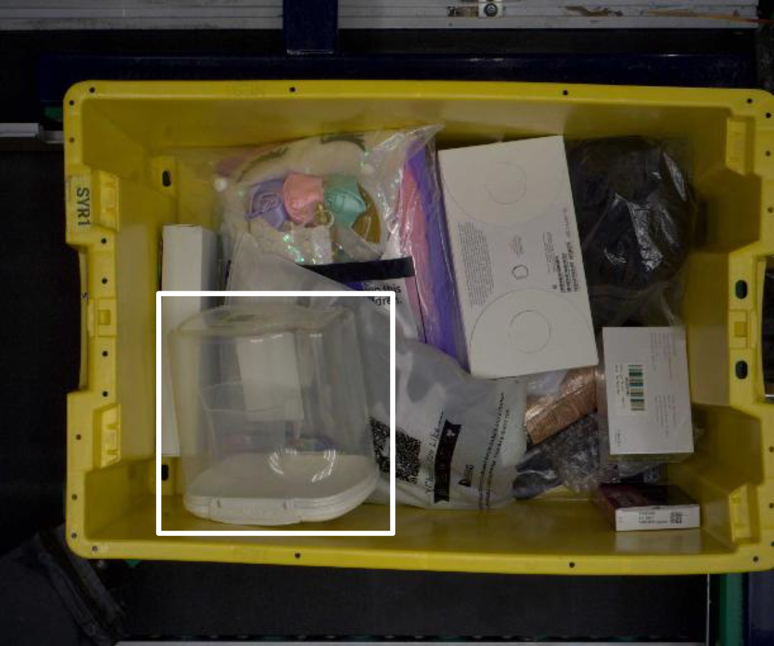}
\caption{Transparent container}
\end{subfigure}\\
\caption{Example pick scenes from the open-set item manipulation task. Some difficult items are highlighted in white boxes.}
\label{fig:more-item-examples}
\end{figure*}

\subsection{Datasets}

We have access to two datasets for training models for item pick success prediction: the first dataset, the \textit{standard} dataset, contains 343K multimodal images of items in unstructured clutter for pretraining the MultiMAE. It also has nearly 275K training examples which contains both multimodal images along with pick features for the specific pick candidate of an item, for which the grasp was executed in deployment. For the executed grasp, we also have an annotation of pick success or failure. The second dataset, which we will refer to as the \textit{random} dataset, is a much smaller dataset containing the same type of data as the 275K training examples mentioned above. The difference with the standard dataset is that the items were picked according to randomly selected pick candidates. So this dataset contains more picks of items that are partially occluded by other items. We also present results on a dataset obtained for a different domain involving the picking of packages.

\subsection{Nature of items}
\label{nature-of-items}

In Figure \ref{fig:more-item-examples}, we further emphasize the challenging nature of our setting with example pick scene images. 

Items such as the big box in (a) can be hard to manipulate due to its heavy weight; some items are deformable such as the plastic bags in (d), the plush toy in (k); some items are transparent, such as the container in (p); some items have irregular or round shapes, making it hard to find a proper pick point, such as the washer pods containers (e), the detergent bottle (f), glass bottles (g), plastic bucket (h), spatula (m), and tape (n); some items can be difficult to pick for multiple reasons, for example the box in (j) is transparent, has an irregular surface and can potentially be heavy, and the toy box in (o) has both deformable and rigid parts, with some transparent packaging. The protein powder bottle in (l) is both round and heavy. It is also important to note that some items can be damaged if not picked properly, such as the glasses in (j), and the book in (b). Additionally, boxes that contain smaller items could open and plastic packaging can be damaged when picked incorrectly. These examples show that reliably picking from an open set of real world items is a very challenging task.

\captionsetup{labelfont={color=black},textfont={color=black}}

\section{Methodology} 

\subsection{Preliminary Evaluation of Visual Encoders}

\begin{figure}[htb]
\centering
\begin{subfigure}[t]{.99\linewidth}
\centering
\includegraphics[width=\linewidth]{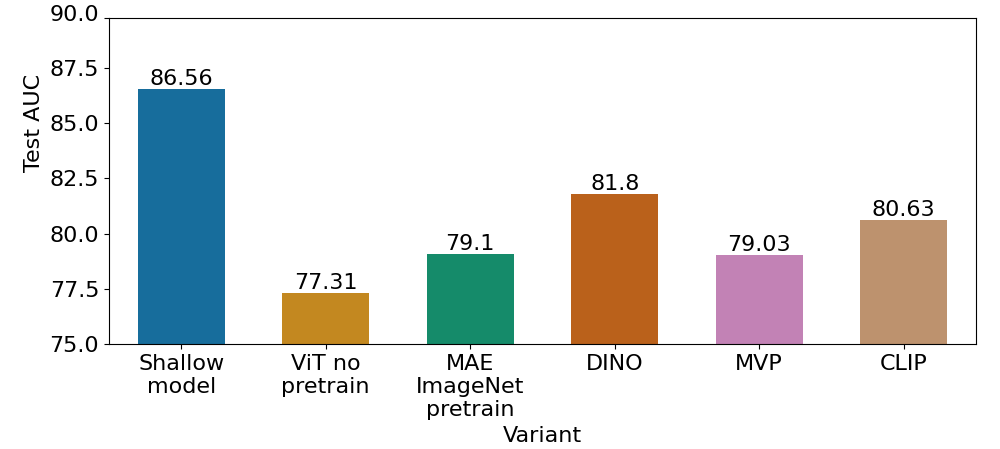}
\end{subfigure}
\caption{Performance of a shallow model compared with pre-trained visual encoders. From left to right: shallow model (XGBoost) baseline that uses expert features; using a ViT model~\cite{DBLP:vit16x16} with random initial weights; ViT model with MAE~\cite{He2021MaskedAA} pretraining on ImageNet; ViT model with DINO~\cite{Caron2021EmergingPI} pretrain; and ViT model with MVP~\cite{xiao2022masked} pretraining. }%
\label{fig:early}
\end{figure}

\begin{figure}[htb]
\centering
\begin{subfigure}[t]{.95\linewidth}
\centering
\includegraphics[width=\linewidth]{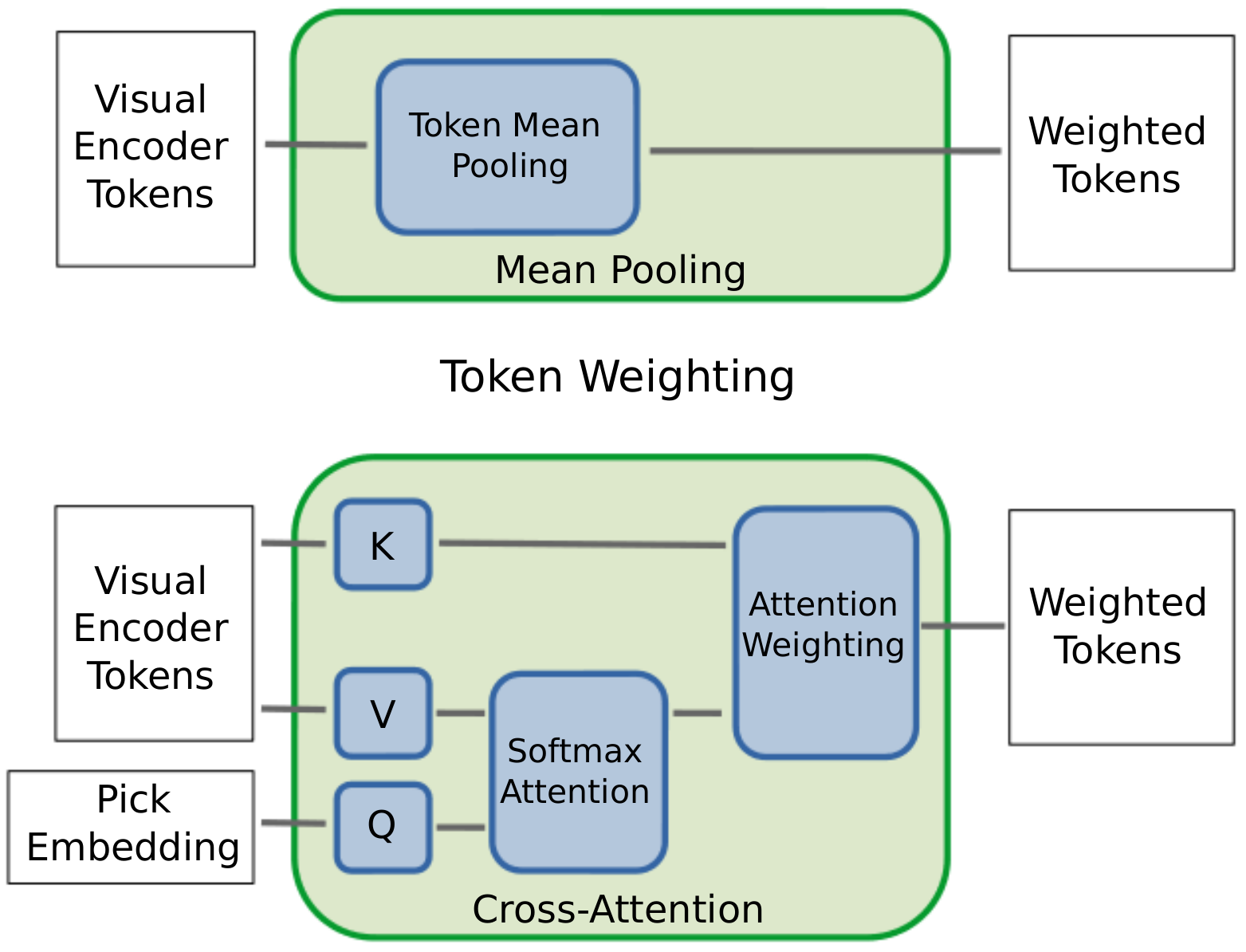}
\end{subfigure}
\caption{Given visual encoder tokens outputted by the MultiMAE encoder,
we implement two types of token weighting: either using a simple mean pooling of all tokens, or weighting via cross-attention. In the latter case, the encoded pick features are also provided as input to the cross-attention block.}
\label{fig:token-weighting}
\end{figure}

Previously, a shallow model with manually engineered features was proposed, which outperformed a deep model with a CNN-based visual encoder~\cite{Li:2023:rss:pickranking}. However, several works in the literature, e.g.,~\cite{xiao2022masked} have proposed to use pretrained visual encoders. We first trained a model similar to that shown in Fig.~\ref{fig:model-schematic}. Instead of the MultiMAE encoder, however, we used a pretrained visual encoder whose weights are frozen, along with token mean pooling for the token weighting (Fig.~\ref{fig:token-weighting}). We train the model on the datasets with annotated pick success (as described above), and unimodal input, i.e., RGB only. Fig.~\ref{fig:early} shows the performance of the shallow model compared to a visual encoder with randomly initialized weights (ViT no pretrain), and various visual encoders pretrained on generic datasets. Results show that using a frozen generic encoder clearly achieves higher performance over an encoder with randomly initialized weights. Nevertheless, the shallow model with expert-engineered features significantly outperforms the pretrained visual encoder models.

\subsection{Demonstrated Strategy}

To close the performance gap with the shallow model, it is evident that pretraining with generic data by itself is not sufficient. We propose several improvements over these models with pretrained visual encoders. Again referring to Fig.~\ref{fig:model-schematic}), we propose the use of a MultiMAE for visual encoding to exploit the availability of multimodal image input. Each new modality will be processed into additional image tokens with a different input adapter. Furthermore, instead of mean pooling, the pick features are input to a cross-attention layer for the Token Weighting. The cross-attention layer aims to learn to relate the encoded pick features over the tokens. Finally, we propose to perform two stages of training: first we train the MultiMAE in a pretraining stage with pixel reconstruction objectives, then we train the entire model in a second stage with the pick success prediction objective. During the second stage we update the weights of the MultiMAE while training. We refer to this second stage as the finetuning stage. Before presenting results and our ablation study, we will first describe the pretraining and finetuning stages in more detail.

\subsection{Pretraining the MultiModal Autoencoder} \label{sec:method:pretraining}

We first pretrain the MultiMAE on 343K examples of deployment data in the standard dataset. The nature of the multimodal data is similar to that in the original MultiMAE work \cite{bachmann2022multimae}. During pretraining, the model receives as input RGB, depth and semantic images of the pick scene. We use a similar schedule as in~\cite{bachmann2022multimae} for randomly masking a portion of the tokens from the three modalities, and the MultiMAE learns to reconstruct the pixels in the missing image patches. When pretraining on the deployment dataset, the MultiMAE is initialized with weights from a publicly available version pretrained on ImageNet, then pretrained for 800 epochs.

\begin{figure}[htb]
\centering
\begin{subfigure}[t]{.95\linewidth}
\centering
\includegraphics[width=\linewidth]{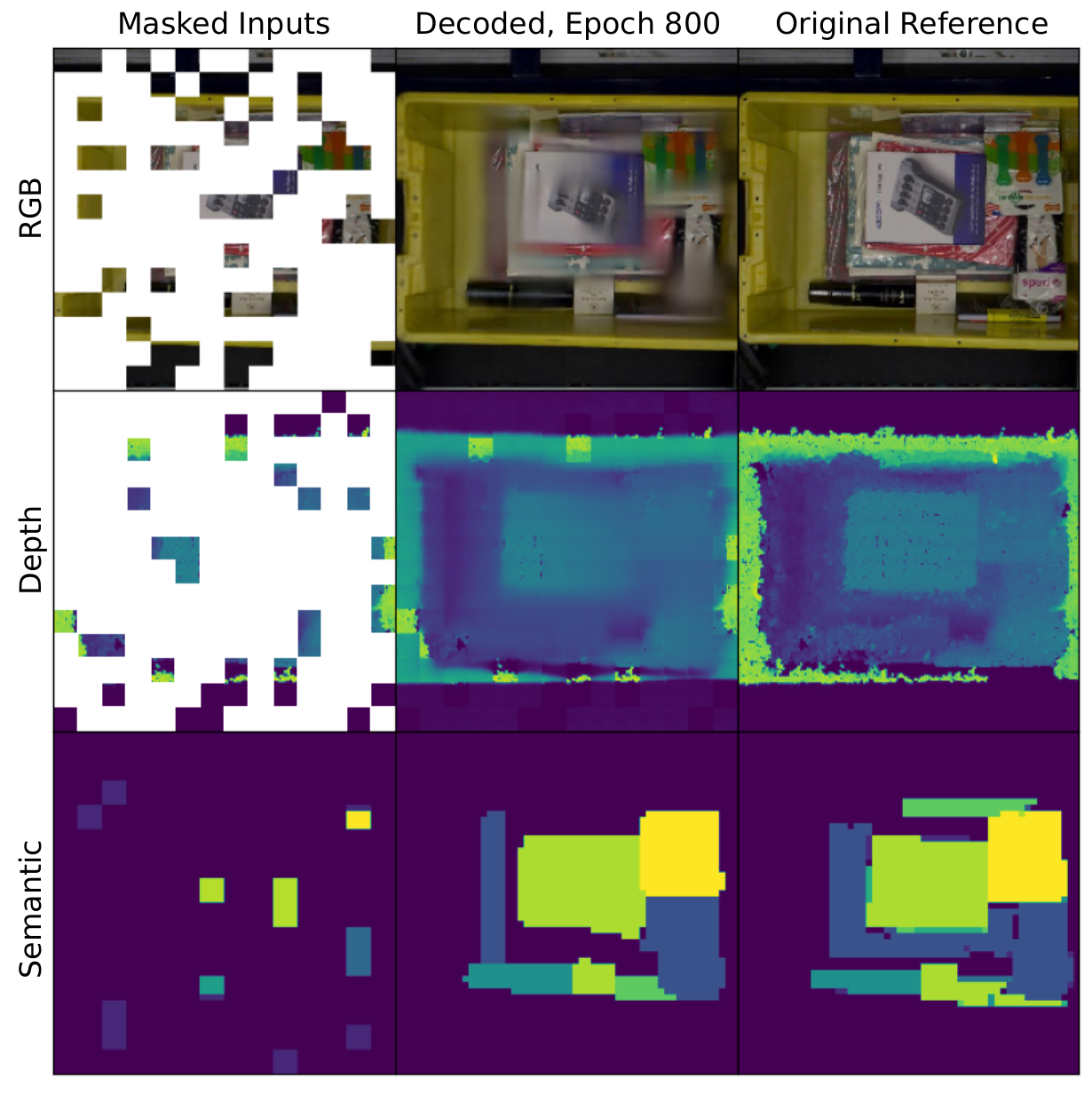}
\end{subfigure}
\caption{Example of MultiMAE reconstruction for all three modalities, i.e., RGB, depth and semantic segmentation. The left column shows the ground truth reference images.}
\label{fig:recon-0}
\end{figure}

At pretraining, the RGB and depth images are resized to $224 \times 224$. 
For the semantic segmentation input images, we first define 9 classes of semantic categories for the items encountered. Instance segmentation images in our dataset, obtained from a previously trained segmentation model, are directly converted into semantic segmentation images. The images are downsampled by a factor of $4 \times$ compared to the RGB and depth data to reduce computation cost~\cite{bachmann2022multimae}. Thus, the resulting dimensions are  $56 \times 56$. At the start of pretraining, we initialize a different class embedding for each semantic class, which is then updated during pretraining. When reconstructing the semantics, MultiMAE will predict a value for each class, i.e., the semantics output has nine channels per pixel. An example of the resulting model's ability to reconstruct all three modalities with randomly masked input is shown in Fig.~\ref{fig:recon-0}. When we visualize the reconstruction, for each pixel, we show the class with the highest prediction value. 

\subsection{Fine-tuning for Pick Success Prediction} \label{sec:method:finetuning}
We next train the MultiMAE along with the additional network components on 275K training examples to perform pick success prediction. We refer to this stage as \textit{finetuning}. We use 69K validation examples to decide when to stop training. The model is then evaluated on 86K test data, which have not been seen during both stages of training. For each split the success to fail ratio is about 11:1. We use a weighted loss to account for this high class imbalance. We also incorporate pick information into the model by using a cross-attention module as shown in Fig.~\ref{fig:token-weighting} bottom. The cross-attention results in a weighted sum of the image tokens based on the pick features.

For simplicity, we discuss the case where we only consider one attention head. We first project the pick embedding into a query token $Q$, and project the encoder tokens (we omit the global \textit{CLS} token) into value tokens $V$ and key tokens $K$. The $V$, $K$, and $Q$ projection layers are simply linear layers and are trained with the pick success objective. Following self-attention~\cite{DBLP:vit16x16}, $Q$ and $K$ are multiplied and then divided by a scale value, followed by a softmax, resulting in an attention weight vector. We then take a weighted sum of the encoder tokens using the attention weight vector. The intuition behind using a cross-attention layer is the fact that we aim to combine information from two entirely different domains: images and pick features. The MultiMAE provides a visual encoding of the multimodal images, which in turn is \textit{queried} with pick features for pick success. In the next section, we will compare cross-attention with the more naive setting of using mean pooling, where each token has the same weight for the attention weight vector. Finally, the weighted sum is then input to a MLP prediction head, together with the projected pick features, to obtain pick success prediction.

Next we will present our experimental results that reflect on the effectiveness of the demonstrated approach.

\section{Experimental Results}

We use the Area Under the Receiver Operating Characteristic Curve (ROC AUC) as our main metric for evaluating the performance. This metric indicates how well the model performs under different decision thresholds and is more suitable for the case where the data is imbalanced~\cite{FAWCETT2006861}. The expert-engineered features are always provided for the shallow model baseline, but not to the demonstrated MultiMAE model. 

\subsection{Experiments}

For our main experimental evaluation, we will adopt the best performing model among the variants of the MultiMAE we have experimented with. We will refer to this as the \textit{demonstrated approach}. The settings for the demonstrated approach are: 
\begin{itemize}
    \item We pretrain on in-domain RGB (R), depth (D) and semantic segmentation (S) images (Pretrain R-D-S);
    \item For the finetuning stage we use RGB and depth images (Finetune R-D), along with an image of the pick location.
    \item The token weighting is done using cross-attention.
    \item An additional pick location image is added to further boost performance.
    \item During finetuning, the MultiMAE encoder weights are updated according to the pick success loss. 
    \item Finally, the input images for finetuning are cropped according to a bounding box with additional padding derived from the target item's segmentation mask. We refer to this as the local crop. The local crops are augmented with random offsets for additional robustness.
\end{itemize} 
Ablation studies that follow will evaluate the importance of the above implementation choices.

\subsubsection{Largely Unoccluded Item Picks - Standard Dataset}

\begin{figure}[htb]
\centering
\begin{subfigure}[t]{.99\linewidth}
\centering
\includegraphics[width=\linewidth]{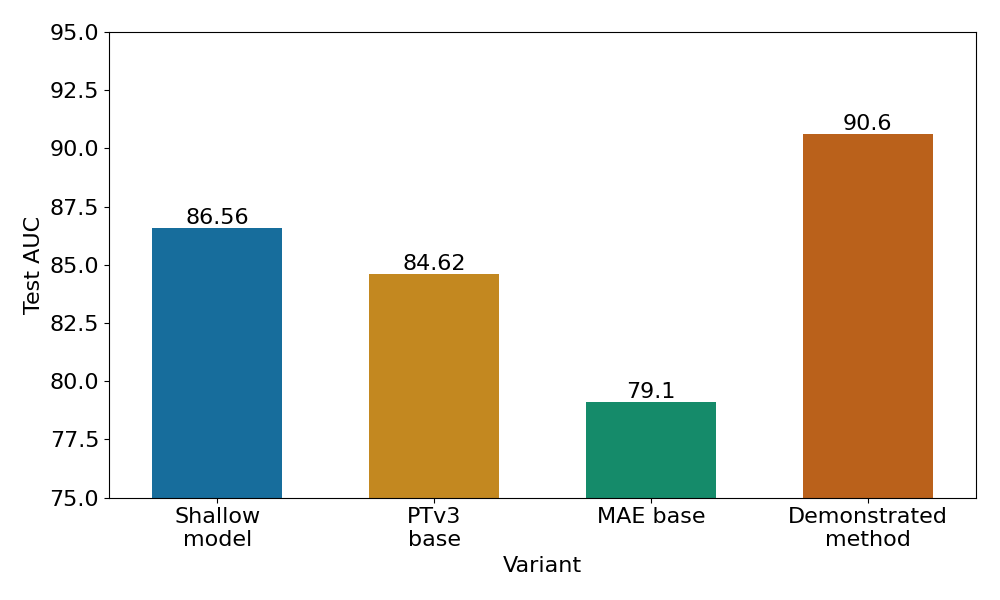}
\end{subfigure}
\vspace{-.2in}
\caption{Performance comparison of the shallow model~\cite{Li:2023:rss:pickranking} (relies on expert features) against a PTv3 baseline, a MAE baseline, and the demonstrated method on a dataset consisting of largely unoccluded item picks. The demonstrated approach outperfoms the shallow model by about 5\% in test AUC.}
\label{fig:main}
\end{figure}

Fig.~\ref{fig:main} shows a performance comparison of the demonstrated MultiMAE approach against the previously best performing method to date on this dataset, i.e., the shallow model~\cite{Li:2023:rss:pickranking}, as well as a baseline version of a (unimodal) mask autoencoder (MAE Base). The demonstrated approach has a significantly higher performance than the MAE Base variant (79.1 vs 90.6). Their differences are studied further in the ablations. The demonstrated approach also outperforms the shallow model by about 4 points without the need to access the expert-engineered features that the shallow model uses.

We also compared against a learn-from-scratch, point-cloud baseline. This baseline adopts the PointTransformerV3 (PTv3) model~\cite{PTv3} as point cloud encoder. Each encoder block receives a point cloud as input and employs: 1) a 3D sparse convolution layer~\cite{spconv2022} to serve as conditional positional embedding, and 2) a self-attention transformer layer on patches created using space filling curves. An action is represented by a set of pick tokens. We use a linear layer to embed a pick feature vector containing the discretized approach angle, wrist rotation, and suction cup activation map to get the feature, and store this for each pick token. The pick token is then used as query in the cross attention and the average location of activated suction cups is used as location for this pick token when performing 3D rotary positional encoding for the cross attention. A linear projection head is then applied to the output of this decoder layer. This baseline results in a test AUC of 84.6, which is stronger than the MAE baseline, but is lower than both the shallow model and the demonstrated MultiMAE model. This again highlights that multimodal pretraining can provide significant benefits in terms of pick success prediction. Note that this result simply shows the scene encoder in the demonstrated approach can also be a point cloud model. Although this paper is focused on 2D multimodal data, we believe proper 3D pretraining as well as combining 3D and 2D modalities can be beneficial, and we plan to further investigate 3D modalities in future work, as discussed in Section \ref{sec:conclusion}. 

\subsubsection{Partially Occluded Item Picks - Random Dataset}
\begin{figure}[htb]
\centering
\begin{subfigure}[t]{\linewidth}
\centering
\includegraphics[width=\linewidth]{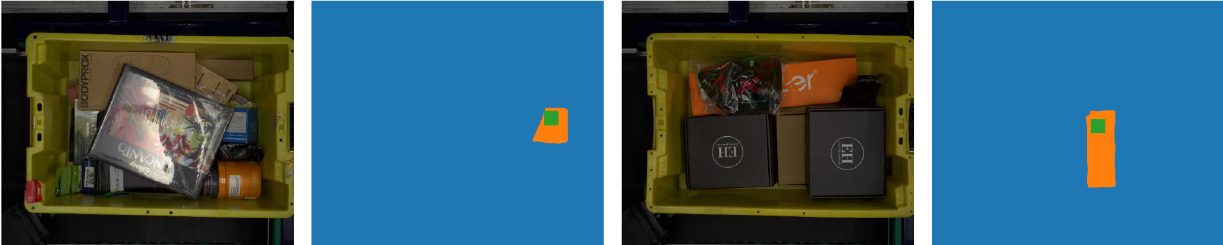}
\end{subfigure}
\caption{Examples of picking partially occluded items. Columns 1\&2, and columns 3\&4 contain an example of an RGB image and corresponding image with the target item mask and pick point location. In both cases, the target item is partially occluded by a larger item at the top of the unstructured pile.}
\label{fig:random-pick-examples}
\end{figure}

In order to evaluate the robustness of the demonstrated approach and learned representation, we also evaluate the demonstrated approach against the shallow model on a random pick dataset. The item configuration distribution of this dataset, with many picks for items that are partially occluded (shown in Fig \ref{fig:random-pick-examples}), is very different from the standard dataset, where we mostly pick unoccluded items. This "random dataset" contains 8,461 training examples, 2,115 validation and 2,644 test examples. The success-failure ratio on each split is approximately 4.4:1.

Table \ref{table:random-dataset} compares the shallow model and the demonstrated approach in three different settings: (a) Pretrain on the standard dataset, finetune on the random dataset, then test on the random dataset (PT: STD, FT: RND, Test: RND); (b) Pretrain on standard, finetune on standard, and then zero-shot test on the random dataset (PT: STD, FT: STD, Test: RND); and (c) Pretrain on standard, finetune on standard first, and then further finetune on the random dataset, then test on the random dataset (PT: STD, FT: BOTH, Test: RND). Note that the shallow model does not have pretraining, and for case (c), the shallow model is finetuned (trained) on the combination of the standard and random datasets. 

The demonstrated approach outperforms the shallow model consistently in all settings. This demonstrates that the representations learned from the demonstrated approach are robust on different item configurations. 

\renewcommand{\arraystretch}{1.25}
\begin{table}[htb]
\small
\caption{Performance comparison between the shallow model and the demonstrated approach on a different dataset where random items that are partially occluded are picked in different settings. The model is pretrained (PT) on standard dataset (STD). And then can be finetuned (FT) on the random dataset (RND), or first finetuned on the standard dataset then on the random dataset (BOTH). Then the model is tested on the random dataset.}
\label{table:random-dataset}
\begin{center}
\begin{tabular}{ c c c c | c }   
\textbf{Model} & \textbf{PT} & \textbf{FT} & \textbf{Inference} & \textbf{Performance}\\
\hline
Shallow  & - & RND & RND & 82.92\\
Demonstrated & STD & RND & RND & \textbf{86.28}\\
\hline
Shallow & - & STD & RND & 83.58\\
Demonstrated & STD & STD & RND & \textbf{85.87}\\
\hline
Shallow & - & BOTH & RND & 85.16\\
Demonstrated & STD & BOTH & RND & \textbf{88.06}\\
\end{tabular}
\end{center}
\end{table}

\subsubsection{Package Picking Dataset}
\begin{figure}[htb]
\centering
\begin{subfigure}[t]{.99\linewidth}
\centering
\includegraphics[width=\linewidth]{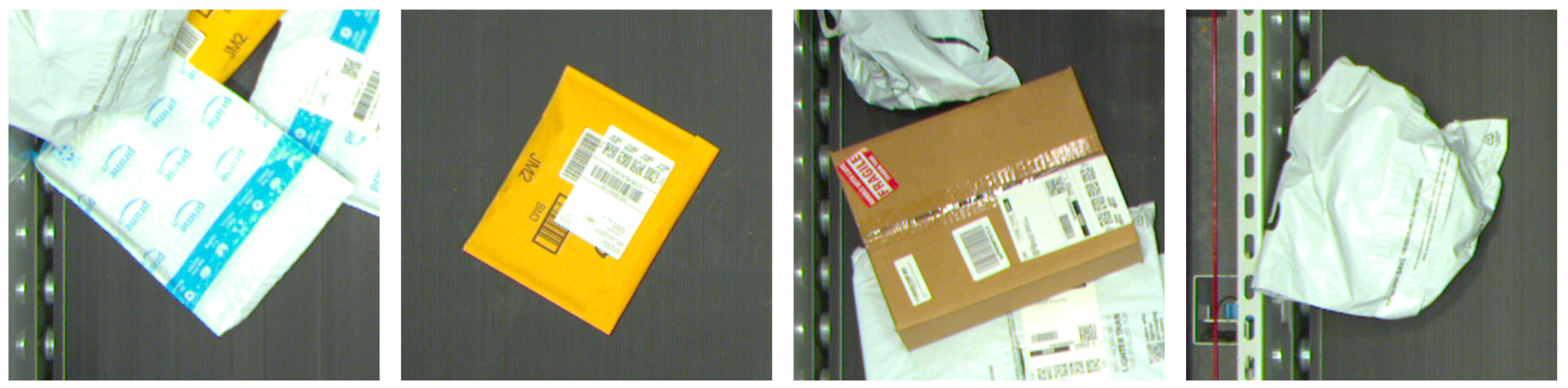}
\end{subfigure}
\caption{Example pick scenes with packages. There is less variety in object appearance relative to the domain of item picking, but the level of clutter can still be challenging.}
\label{fig:robin-examples}
\end{figure}

We also investigate whether the demonstrated approach can work on a different pick scene containing packages rather than items. Here we pretrain and finetune on the multimodal data derived from the package manipulation task described in~\cite{Li:2023:rss:pickranking}. This is quite different from the item manipulation setting, and the majority of the items to pick are packages such as boxes, bags and envelopes, as shown in Fig.~\ref{fig:robin-examples}. Here we consider a package picking dataset with 100K training, 20K validation and 20K test examples, each with 2:1 success-failure ratio.

Table \ref{table:package-dataset} shows when we perform multimodal pretraining and finetuning on this package picking dataset, we can also outperform the shallow model that relies on expert features, and obtain the best performance. This demonstrates the demonstrated approach also works with a different pick scene with a different item distribution.

\renewcommand{\arraystretch}{1.25}
\begin{table}[htb]
\small
\caption{Package picking dataset experiment. Results show that the demonstrated approach can also outperform the shallow model for this different setting where the majority of the items to pick are packages such as boxes, bags and envelopes.}
\label{table:package-dataset}
\begin{center}
\begin{tabular}{ l | l  l } 
\textbf{Package Picking Dataset Experiment} & \textbf{Performance} \\
\hline
Shallow Model & 86.50\\
No Pretrain & 84.54\\
Generic RGB Pretrain, Frozen & 86.99\\
In-domain Multimodal Pretrain, Finetune & \textbf{88.28}\\
Item Picking Data Pretrain & 87.40 \\
Item Picking Data Pretrain, Frozen & 85.43 \\
\end{tabular}
\end{center}
\end{table}

\subsection{Ablation Studies}

We present a series of ablations to better understand what the most important factors are for the performance of the demonstrated method, and how performance differs with variations of data, input modalities and other settings.

\subsubsection{Effect of Visual Modalities}
Table \ref{table:modality-ablation} shows the effect of having different combinations of the visual modalities, i.e., RGB (R), depth (D) and semantics (S), at the pretraining and finetuning stages. The results show that pretraining with more modalities can bring performance gain even when finetuning with RGB only. When pretrained with all three modalities, having depth and semantics as additional input at the finetuning stage can also further improve performance. When only a single modality is used at finetuning, RGB has the best performance, followed by depth and semantics. Note that for our demonstrated approach, we do not use semantics during finetuning, since further adding semantics will increase latency while providing only minimal performance improvement.

\renewcommand{\arraystretch}{1.25}
\begin{table}[htb]
\small
\caption{Visual modality ablation. Comparison of performance for pretraining and finetuning with different modalities. The modalities are RGB (R), depth (D), and semantic segmentation (S) images. The effect is measured with respect to the default setting, denoted with **. The default setting is also described in Figure \ref{fig:model-schematic}. The improvement of using all three modalities for both pretraining and finetuning (last row) is minimal. The best performance is highlighted in bold.}
\label{table:modality-ablation}
\begin{center}
\begin{tabular}{ p{2.5cm} l | l | l }   
\multicolumn{2}{l|}{\textbf{Visual Modality Ablation}} & \textbf{Performance} & \textbf{Effect} \\
\hline
Pretrain: R & \hspace{-0.75cm} Finetune: R & 87.35 & -3.24 \\
Pretrain: R-D & \hspace{-0.75cm} Finetune: R & 89.11 & -1.49 \\
Pretrain: R-S & \hspace{-0.75cm} Finetune: R & 89.93 & -0.67 \\
Pretrain: R-D-S & \hspace{-0.75cm} Finetune: R & 90.05 & -0.55 \\
Pretrain: R-D-S & \hspace{-0.75cm} Finetune: D & 87.86 & -2.74 \\
Pretrain: R-D-S & \hspace{-0.75cm} Finetune: S & 85.25 & -5.35 \\
Pretrain: R-D-S & \hspace{-0.75cm} Finetune: R-D** & 90.60 & 0.00 \\
Pretrain: R-D-S & \hspace{-0.75cm} Finetune: R-S & 90.14 & -0.46 \\
Pretrain: R-D-S & \hspace{-0.75cm} Finetune: R-D-S & \textbf{90.76} & 0.17 \\
\end{tabular}
\end{center}
\end{table}


\subsubsection{Effect of In-Domain Pretraining}
Many popular works in the literature advocate for the use of frozen visual representations that are pretrained on large generic datasets \cite{xiao2022masked, parisi2022unsurprising, nair2022r3m}. This is reasonable for tasks where only a small amount of in-domain data is available. Table \ref{table:pretrain-domain-ablation} shows how pick prediction success performance can be affected heavily by the pretraining dataset. Although pretraining on generic datasets (ImageNet) with either RGB only or all three modalities will improve performance over not pretraining the visual encoder at all, pretraining on in-domain data with all three modalities can further boost performance (row 4 in Table~\ref{table:pretrain-domain-ablation}).

\begin{table}[htb]
\small
\caption{Pretrain domain ablation. Comparison of performance for different pretraining settings. Generic: pretrain on ImageNet, in-domain: pretrain on deployment data. Pretraining with RGB, depth, and semantic segmentation on in-domain data (indicated by **) achieves highest performance.}
\label{table:pretrain-domain-ablation}
\begin{center}
\begin{tabular}{ l | l | l } 
\textbf{Pretrain Domain Ablation} & \textbf{Performance} & \textbf{Effect} \\
\hline
No pretrain & 80.44 & -10.16 \\
Pretrain: R generic & 81.84 & -8.76 \\
Pretrain: R-D-S generic & 84.43 & -6.17 \\
Pretrain: R-D-S in-domain ** & \textbf{90.60} & 0.00 \\
\end{tabular}
\end{center}
\end{table}

\subsubsection{Effect of Local Crop Sizes}
\begin{figure}[htb]
\captionsetup[subfigure]{justification=centering}
\centering
\begin{subfigure}[t]{.24\linewidth}
\centering
\includegraphics[width=\linewidth]{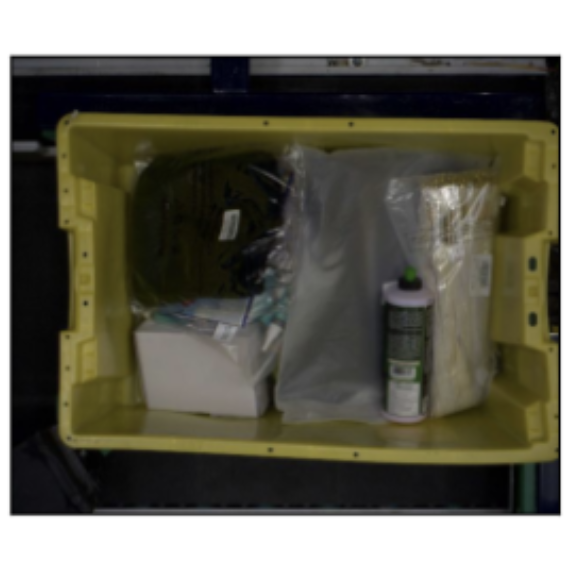}
\caption{Global (89.32)}
\end{subfigure}
\begin{subfigure}[t]{.24\linewidth}
\centering
\includegraphics[width=\linewidth]{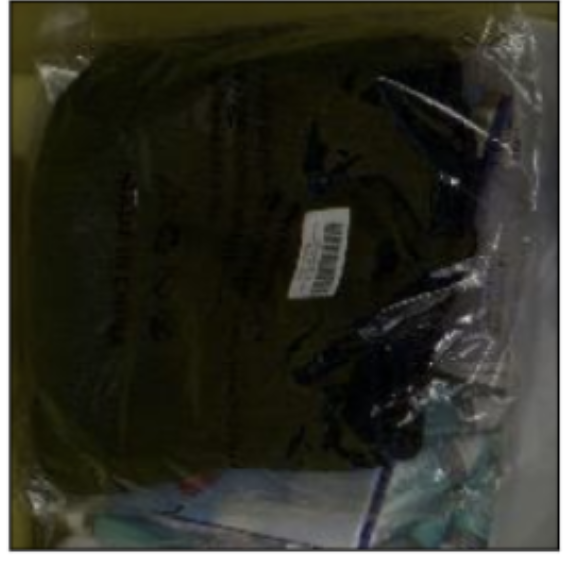}
\caption{Local+0 (89.78)}
\end{subfigure}
\begin{subfigure}[t]{.24\linewidth}
\centering
\includegraphics[width=\linewidth]{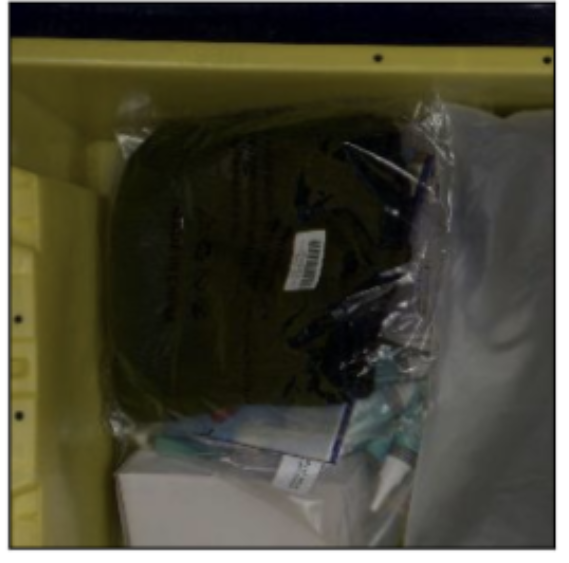}
\caption{Local+50 (90.60)}
\end{subfigure}
\begin{subfigure}[t]{.24\linewidth}
\centering
\includegraphics[width=\linewidth]{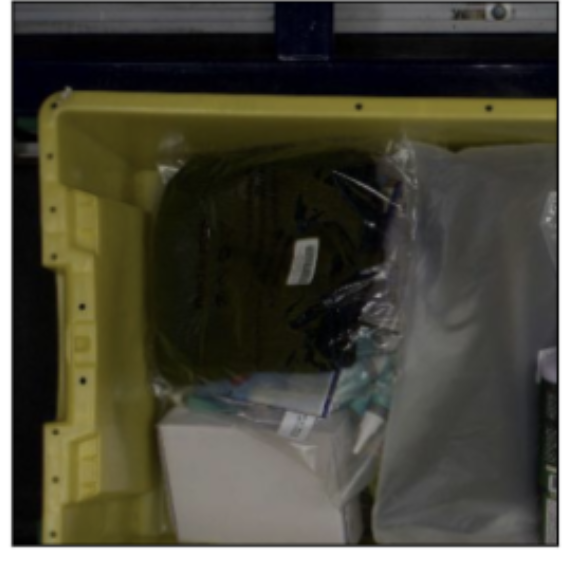}
\caption{Local+100 (90.42)}
\end{subfigure}
\caption{Global image vs local crop centered around the target item with different padding values. This is also described in Figure \ref{fig:model-schematic}. Performance (in parenthesis) with local crops is better compared to the global image. A padding of 50 is the best.
gives the best performance.
}
\label{fig:abl3-illustration-of-different-offsets}
\end{figure}

For the visual input at the finetuning stage, we can either use the image of the entire pick scene, or we can crop out a portion of the image centered around the target item. The size of the crop is determined by the segment bounding box and a padding value. We note that the model is pretrained on both global and local crop due to random crop augmentation following \cite{bachmann2022multimae}. A padding of 0 means the crop is tight around the target item segment, larger values for the padding will  include more surrounding information. When we have multiple visual modalities, we crop all of them the same way. Fig.~\ref{fig:abl3-illustration-of-different-offsets} shows the global scene image and the same image with different local crop sizes. The performance we get when using them as input is shown directly below each image. A local crop is better than a global image, a padding value of 50 gives the best score.

\subsubsection{Different Ways to Incorporate Pick Features}
The pick success prediction model needs to integrate information from very different domains: visual modalities and encoded pick features, which contain 3D pose and other information relevant to the pick, e.g., activated suction cups of the multi-suction cup end effector. How we combine the images and the pick information can affect the performance. In this ablation we study three different ways to incorporate the pick features: (1) use a cross-attention module for learned token weighting; (2) use a local crop of the input images, centered around the target item, instead of the entire (global) image; (3) mark the pick point on another 2D image, and use it as an additional visual input modality. Table \ref{table:pick-location-incorporation} shows how different ways of incorporating pick location affect performance.

\begin{table}[htb]
\small
\caption{Pick incorporation ablation. Comparison of performance for different ways to incorporate pick location information. Here ``cross-attn'' means use cross attention to compute weighted tokens; ``pick loc image'' means marking the pick point on a 2D image, and using it as an additional visual input modality. The effect is measured with respect to the default setting, denoted with **.}
\label{table:pick-location-incorporation}
\begin{center}
\begin{tabular}{ l | l | l } 
\textbf{Pick Incorporation Ablation} & \textbf{Performance} & \textbf{Effect} \\
\hline
Global mean pool w/o pick loc image & 85.76 & -4.84 \\
Global cross-attn w/o pick loc image & \textbf{88.55} & -2.05 \\
\hline
Local mean pool w/o pick loc image& 89.52 & -1.08 \\
Local cross-attn w/o pick loc image & \textbf{89.76} & -0.84 \\
\hline
Local mean pool w/ pick loc image & 90.51 & -0.09 \\
Local cross-attn w/ pick loc image ** & \textbf{90.60} & 0.00 \\
\end{tabular}
\end{center}
\end{table}

\begin{figure}[htb]
\centering
\begin{subfigure}[t]{.99\linewidth}
\centering
\includegraphics[width=\linewidth]{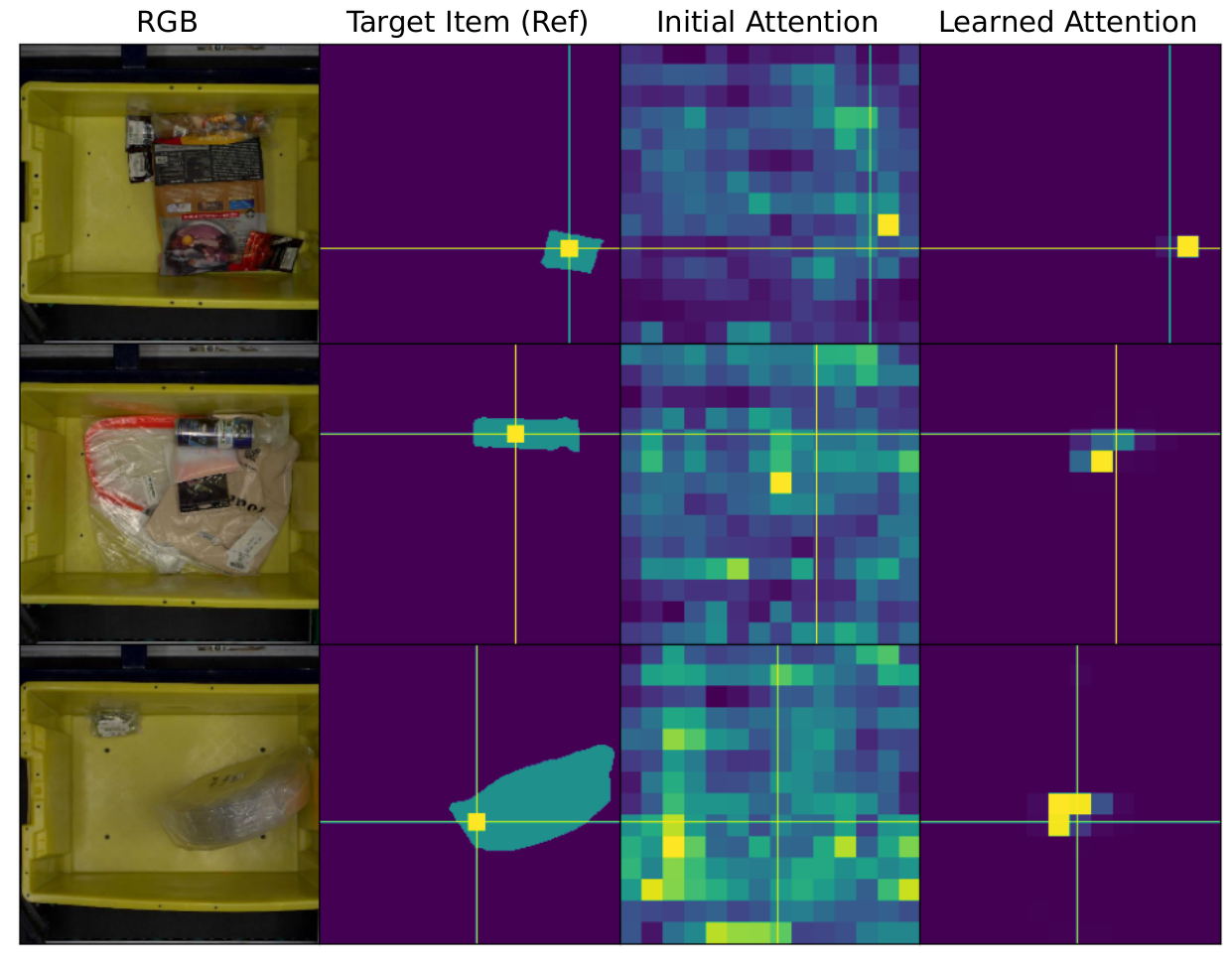}
\end{subfigure}
\caption{Different examples of visualization of the learned attention. In this particular case, only the RGB and the pick coordinates are provided to the model. The target item mask and pick point in the second column are only for reference. The model is able to learn to pay attention to regions near the target item and pick point. }
\label{fig:abl5-cross-attention-visualize}
\end{figure}

With global scene multimodal images, without providing a pick location image, and using mean pooling of the encoded image tokens as the weighting (refer to Fig.~\ref{fig:token-weighting}), it is hard for the model to identify the target item when there are multiple items in the scene (row 1 in Table~\ref{table:pick-location-incorporation}). Adding the cross attention module allows the model to associate the pick location information of the pick features with the pick point location and target item in the image. Fig.~\ref{fig:abl5-cross-attention-visualize} shows evidence that in this case, the model is able to learn (through the pick success prediction loss) which image tokens to pay attention to, which improves the performance (row 2 in Table~\ref{table:pick-location-incorporation}.

The right two columns of Fig.~\ref{fig:abl5-cross-attention-visualize} show the model initially has random attention, but learns to focus more on the target item and image patches near the pick point. We also visualize the pick point as a small square in the second column, together with the target item segmentation mask. The crosshair in the three columns on the right is only used for visualization purpose in the visualized attention map. Note that in this particular experiment, the target item mask and the pick point square are only provided as a reference and are not available to the model during training or inference.

If we switch to using a local crop image centered around the target item, the model can more easily combine the pick features with the encoded images, leading to performance improvement (row 3, Table~\ref{table:pick-location-incorporation}). However, notice that the positive effect of cross attention is reduced in the local input setting (compare rows 3 and 4).

\begin{figure}[htb]
\centering
\begin{subfigure}[t]{.99\linewidth}
\centering
\includegraphics[width=\linewidth]{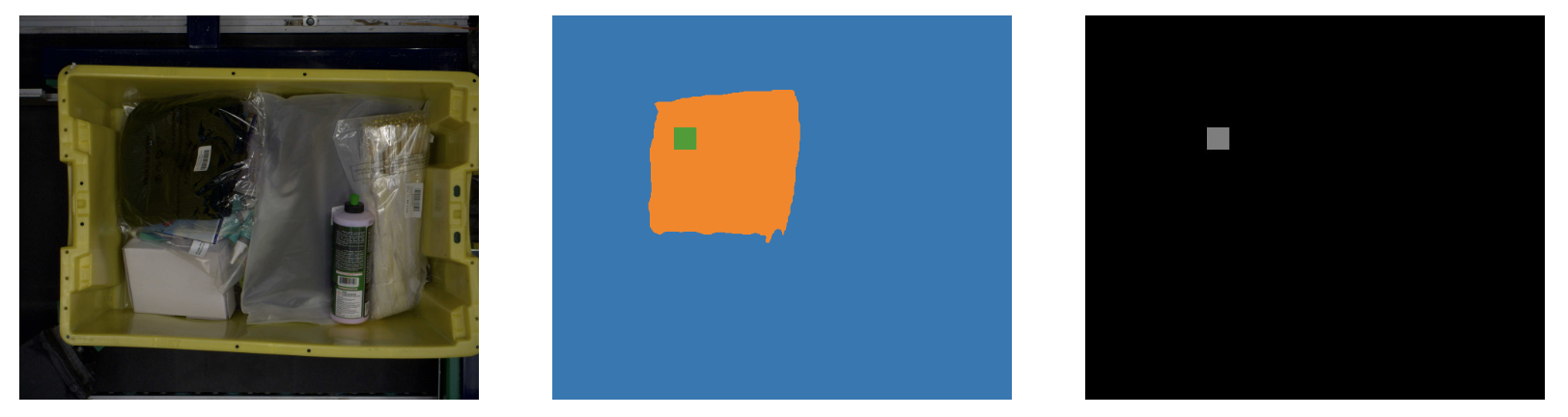}
\end{subfigure}
\caption{Left: RGB input image. Middle: pick location and target item mask for our reference. Right: the pick location image which is used as an additional input.}
\label{fig:rgb-and-pick-point}
\end{figure}

Inspired by recent work that shows explicitly marking object location with a single pixel can help manipulation~\cite{stone2023open}, we also tried providing the pick location explicitly as an additional 2D image, and found it can further boost performance. Here the pick location is marked by a square of pixels on a $224 \times 224$ single channel image, as shown in Fig.~\ref{fig:rgb-and-pick-point}. The last two rows in Table \ref{table:pick-location-incorporation} show this can bring performance to 90.60 (Local cross attention + pick loc image). Again notice that the effect of cross attention is weaker when we use a local crop together with the pick location image.

These results show that using cross attention, local crop and pick location image can all improve performance, and their effects are partly overlapping. In practice, the best input setting can depend on the use case, e.g. when evaluating large amounts of pick candidates in a scene, using a global image with cross attention and without pick location image can give the lowest overhead. So we include all three components in our demonstrated approach for more robustness.

\subsubsection{Effect of Finetuning the Visual Encoder}
We found that finetuning the visual encoder with pick success prediction leads to improved performance compared to freezing it:  87.89 vs 90.6 as shown in Table \ref{table:abl-encoder-finetune}. While it can be good to use a frozen visual encoder in a small data setting, this result shows we should further finetune when a large dataset is available.

\begin{table}[htb]
\small
\caption{Performance of the demonstrated method with or without encoder finetuning at finetuning stage.}
\label{table:abl-encoder-finetune}
\begin{center}
\begin{tabular}{ l | l | l }
\textbf{Encoder Finetuning} & \textbf{Performance} & \textbf{Effect} \\
\hline
Frozen visual encoder & 87.89 & -2.71 \\
Finetuned visual encoder & \textbf{90.60} & 0.00 \\
\end{tabular}
\end{center}
\end{table}

\subsubsection{Effect of Data Augmentation}
In our experiments we also explored whether using data augmentation during finetuning helps performance. The data augmentation is a random shift of the local crop. The crop is done on images of shape $512 \times 612$ (done before they are resized to $224 \times 224$). The center of the crop is initially the center of the item segment. When data augmentation is used, the center can be shifted to four directions with a random value (-25 to 25 pixels). We also use a random crop padding of up to 150 pixels (in 50 pixel increments) to change the size of the crop. We then ensure the target item is always in view, so if the crop region is shifted too much that the item is missing in the view, then the crop region is enlarged to include the target item. The same crop is applied to all visual inputs (RGB, depth, semantics, and pick location image). Augmentation is only applied for training examples and not used during validation or testing. Table \ref{table:abl-data-aug} shows that although our dataset is fairly large, data augmentation still can provide marginal improvement.

\begin{table}[htb]
\small
\caption{Performance of the demonstrated method with or without data augmentation.}
\label{table:abl-data-aug}
\begin{center}
\begin{tabular}{ l | l | l }
\textbf{Data Augmentation Ablation} & \textbf{Performance} & \textbf{Effect} \\
\hline
Without data augmentation & 90.40 & -0.20 \\
With data augmentation & \textbf{90.60} & 0.00 \\
\end{tabular}
\end{center}
\end{table}
\subsubsection{Effect of Pretraining Epochs}
Table \ref{table:abl-pretrain-epoch} shows how much performance changes as we pretrain for a larger number of epochs on in-domain data. We always start in-domain pretraining with generic weights pretrained on ImageNet containing three modalities\cite{bachmann2022multimae}. On row 1, "0 epochs" means we use the generic pretrained weights without any in-domain pretraining, and then directly go into the finetuning stage. The results show that performance improvement gains are largest in earlier epochs of training, and after pretraining for 200 epochs the performance starts to improve more slowly. Nevertheless, the best performance is obtained with the most epochs, which is 800 in our experiments.

\begin{table}[htb]
\small
\caption{Performance of the demonstrated method with different number of pretraining epochs on in-domain data. Numbers after the dashed line indicate higher performance compared to the shallow model. }
\label{table:abl-pretrain-epoch}
\begin{center}
\begin{tabular}{ l | l | l }
\textbf{Pretrain Epoch Ablation} & \textbf{Performance} & \textbf{Effect} \\
\hline
0 epochs & 84.43 & -6.17 \\
\hdashline
20 epochs & 87.48 & -3.12 \\
100 epochs & 90.08 & -0.52 \\
200 epochs & 90.53 & -0.07 \\
400 epochs & 90.51 & -0.09 \\
800 epochs & \textbf{90.60} & 0.00 \\
\end{tabular}
\end{center}
\end{table}

\subsubsection{Effect of Pretraining Data Ratio}
Table \ref{table:abl-pretrain-ratio} shows how much performance changes as we pretrain on different ratios of the in-domain dataset. These results show that even when pretrained on 1\% of the in-domain data (3.4K), there is already a significant benefit, and the performance can be further improved when we have more in-domain data for pretraining. 

\begin{table}[htb]
\small
\caption{Performance of the demonstrated method when pretrained on different data ratios. By default, we pretrain on 100\% of the data (343K). Numbers after the dashed line indicate higher performance than the shallow model. With just 1\% of in-domain the model can outperfom the shallow model.}
\label{table:abl-pretrain-ratio}
\begin{center}
\begin{tabular}{ l | l | l }
\textbf{Pretrain Ratio Ablation} & \textbf{Performance} & \textbf{Effect} \\
\hline
0\% & 84.43 & -6.17 \\
\hdashline
1\% (3.4K) & 87.90 & -2.70 \\
10\% (34K) & 89.46 & -1.14 \\
25\% (86K) & 90.14 & -0.46 \\
50\% (172K) & 90.38 & -0.22 \\
100\% (343K) & 90.60 & 0.00 \\
\end{tabular}
\end{center}
\end{table}

\subsubsection{Alternative Representation Learning Methods with In-domain Data}
\label{abl-alternative-rl-methods}

To further understand how other representation learning methods perform with in-domain pretraining, we conducted experiments to compare the performance of MultiMAE, DINO and MOCO-v3 when they are pretrained with in-domain data, then finetuned under the same setting. We use MOCO-v3 instead of CLIP since our dataset does not have text annotation. 

For a fair comparison, we first experiment with 4 settings, where we pretrain on in-domain data for 100 epoches using MAE, MultiMAE, DINO, MOCO-v3, respectively. For each setting we then use the exact same finetune set up, with exactly the same hyperparameters and model architecture as the demonstrated method, and using RGB as the only visual input. Since MultiMAE is able to leverage additional visual modalities, we additionally add a setting where we pretrain with MultiMAE for 100 epoches, then finetune with RGB, depth image and pick location image (which is the default setting we used in the demonstrated method). These experiments help us better understand the impact of in-domain pretraining and multi-modal learning, and whether other representation learning methods with a single modality can achieve the performance of multi-modal pretraining and finetuning. 

The results are summarized in Table \ref{table:in-domain-methods-ablation}. The results show that with in-domain pretraining, MAE, DINO and MOCO-v3 achieve similar results, with MOCO-v3 slightly better than MAE ($85.32 > 84.85$), and DINO slightly better than MOCO-v3  ($85.88 > 85.32$). All three of them are slightly weaker than our shallow baseline (86.56). However, when we use MultiMAE for multi-modality pretraining, even when we only use RGB as the visual modality at finetuning, we see the performance improves to 87.97, outperforming the baseline. 

When we add more visual modalities to the finetuning stage (RGB, depth and pick location image, same as the default setting in our demonstrated method), we see performance further improves to 90.08. These results are consistent with the other ablations and show that both in-domain pretrain and multi-modal pretrain + finetuning are critical to the final superior performance.

\renewcommand{\arraystretch}{1.25}
\begin{table}[htb]
\small
\caption{Comparing different representation learning methods with in-domain pretrain.}
\label{table:in-domain-methods-ablation}
\begin{center}
\begin{tabular}{ l | l | l }   
\textbf{Pretrain Method} & \textbf{Finetune Modality} &  \textbf{Performance}\\
\hline
MultiMAE & RGB, Depth, Pick location & \textbf{90.08} \\
MultiMAE & RGB & 87.97 \\
DINO & RGB & 85.88 \\
MOCO-v3 & RGB & 85.32 \\
MAE & RGB & 84.85 \\
\end{tabular}
\end{center}
\end{table}

\textbf{\textit{Takeaways from Ablations:}}
A number of insights are revealed from these ablations:
\begin{enumerate}
    \item In-domain and multimodal pretraining gives the largest performance boost, the benefit from multimodal pretrain persists even if only RGB is used at finetuning. 
    \item Using a local crop around the target item with some surrounding information gives best performance.
    \item Cross-attention allows the model to focus on the target item, while a pick location image gives further boost. 
    \item Further finetuning the encoder with pick success prediction loss is better than freezing weights. 
    \item Data augmentation helps even with large-scale data. 
    \item Pretraining for longer epochs gives better performance with diminishing returns. 
    \item Significant gains can be achieved by just performing in-domain multimodal pretraining on 1\% of the available training data.
\end{enumerate}

\subsection{Impact of the Demonstrated Approach}

\subsubsection{Confusion Matrix}
\label{section-cm}
\begin{figure}[htb]
\captionsetup[subfigure]{justification=centering}
\centering
\begin{subfigure}[t]{.49\linewidth}
\centering
\includegraphics[width=\linewidth]{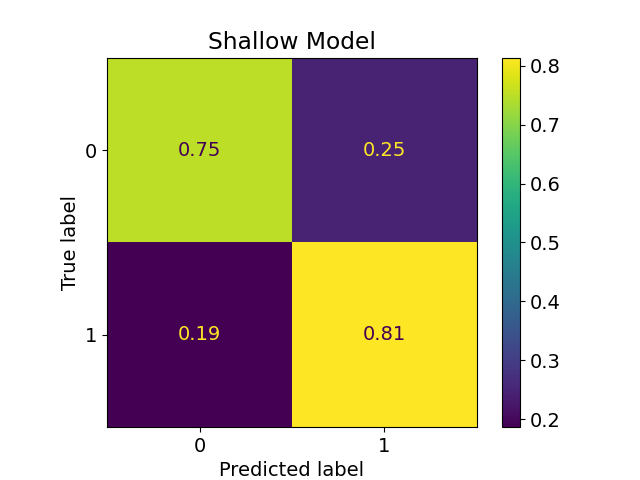}
\end{subfigure}
\begin{subfigure}[t]{.49\linewidth}
\centering
\includegraphics[width=\linewidth]{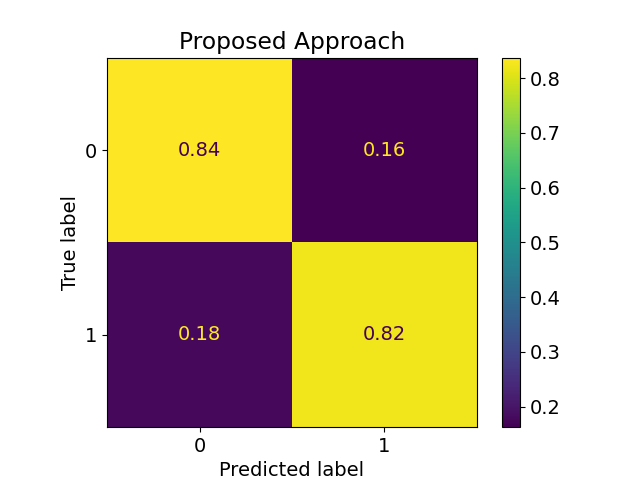}
\end{subfigure}
\caption{Confusion matrices on the test set of the standard dataset, for the shallow model and the demonstrated approach.}
\label{fig:cm}
\end{figure}

The confusion matrix in Figure \ref{fig:cm} shows that the demonstrated approach gives 9\% lower false positive rate and 1\% lower false negative rate, compared to the baseline. 

It is important to note that each failed pick in the real world can have a cost depending on the item type and the type of failure. For example, if a failed pick caused the item to remain in the tote without damage, then the model can try to pick it up again, and the cost is relatively small. 

However, there are other types of pick failure. If the picked item dropped to the ground, it may require human intervention to remove it or it might interfere with other robotic units nearby. If multiple items are picked, it will lead to further error in downstream tasks in the automated system. And if the item is damaged, then it has to be removed and replaced. These failure cases can have a very high cost.

\subsubsection{Real World Test Deployment}
\label{section-cm}
In a real world test deployment of the demonstrated approach, we used the demonstrated approach to perform about 17K pick attempts. Over the course of one week, compared with our baseline method, this approach can potentially reduce multipick by 2\%, mispick by 38\% and amnesty by 41\%, demonstrating its potential for significant cost reduction when deployed at the entire fleet.

In summary, improving the metric by a few percent might not be a big deal in a lab setting, but its impact can be very significant for a large-scale real world system.

\section{Conclusion} \label{sec:conclusion}
This paper focuses on the highly challenging, and less explored large-scale real-world robotic item picking setting with multi-suction grippers. It proposes a way to pretrain and finetune a visual encoder to learn useful multimodal visual representations that are better than expert-engineered features. The demonstrated approach can significantly and consistently outperform the previously best performing shallow model on the task of pick success prediction over three datasets with different item configuration, pick scene and object type. It also outperforms a learn-from-scratch point-cloud encoder alternative. 
Extensive ablations further highlight the critical technical components that lead to this strong performance, and reveal a number of useful insights.

The promising results in this paper open up a number of future directions. One direction is to further incorporate more modalities, such as text, which can allow the learned representations to be better used in other tasks in the same domain \cite{nair2022r3m, karamcheti2023language}, such as damage prediction and targeted picking. It is also interesting to further investigate how to effectively use 3D data such as point cloud,
or design new pretraining schemes that are tailored towards robotic data and task settings. 

\section{Limitations} \label{sec:limitations}
The limitations of this paper are as follows: 
(1) The learned success prediction model relies on a heuristic pick generator that outputs candidate picks. Developing a learning-based, robust pick generator is an interesting direction that can build on top of the demonstrated approach. (2) The extensive experiments presented in this paper are focused on the setting of a single robotic arm with a multi-suction end-effector, and do not include other embodiments, such as multi-arm settings or other end-effector types, such as pinch gripper or soft deformable gripper.  (3) This paper focuses on the highly-challenging, open-set item picking challenge in a industrial warehouse setting. The results are not tested in other environments, such as a household or hospital setting.


\vspace{20pt}


\section*{Acknowledgments}
We want to thank everyone at Amazon Robotics for providing helpful insights and support in many aspects throughout the research. 
We want to especially thank Azarakhsh Keipour, Mostafa Hussein for providing critical feedback with an early version of the paper. 

\bibliographystyle{plainnat}
\bibliography{references}

\clearpage
\appendix

\section{Additional Experiments}

Additional experiments and details:

\subsubsection{Scaling Trend with Finetune Dataset Size}
\label{sec-scaling}

\begin{figure}[htb]
\captionsetup[subfigure]{justification=centering}
\centering
\begin{subfigure}[t]{.99\linewidth}
\centering
\includegraphics[width=\linewidth]{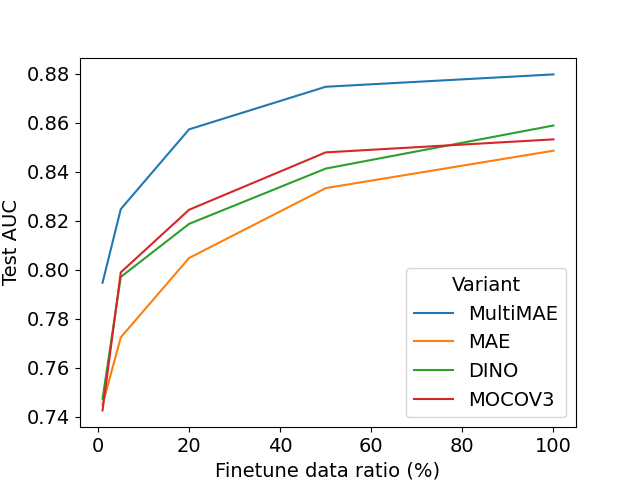}
\end{subfigure}
\caption{Scaling trend of different representation learning methods with respect to finetune data sizes. }
\label{fig:ft-data-ratio}
\end{figure}

In Figure \ref{fig:ft-data-ratio} we show how performance scales with different finetune dataset sizes for the demonstrated approach as well as alternative representation learning methods. For a fair comparison, for each method, we again pretrain for 100 epoches on in-domain data, and keep all finetuning settings and hyperparameters exactly the same. We also use RGB as the only input during finetuning stage, so that MultiMAE does not have an advantage from using multi-modal data at finetune stage. 

The result shows that MAE has slightly weaker performance than DINO and MOCOv3, DINO is slightly weaker than MOCOv3 but start to become better at a larger data size, and MultiMAE consistently outperform the other methods at all finetune data ratios (we tried use 1\%, 5\%, 20\%, 50\% and 100\% of the finetune data). 

Extensive discussions on the scaling behavior of MAE compared to other representation learning methods can also be found in \citet{He2021MaskedAA} and \citet{bachmann2022multimae}. Note that our approach can work with any visual encoders. So we can easily replace MultiMAE with a different and more performant multi-modal representation learning approach in the future.

\subsubsection{Effect of Stacking Modalities}
We also test different representation learning objectives on multi-modal data by channel-wise stacking the R, G, B, D, S inputs.

We conducted experiments where we stack these inputs into 5 channels, and finetune MultiMAE, MAE, DINO and MoCoV3 under the same setting as in Table 10 (in-domain pretraining using standard settings for 100 epochs, then finetune with the stacked image and pick parameters). The results are shown in Table \ref{table:stacking}. MultiMAE is still the best performing model. It is better than MultiMAE with RGB only ($88.52 > 87.97$), but worse than the recommended MultiMAE with 3 separate visual modalities ($88.52 < 90.8$). This shows stacking visual modalities is better than using a single modality, but weaker than using each modality separately. The stacking does make inference faster (from 687 ms to 194 ms, batch of 128). 

\begin{table}[h]
\centering
\caption{Performance comparison when stacking RGB, depth and semantic inputs.}
\label{table:stacking}
\begin{tabular}{l|cccc}
Objective: $\ \ $ & $\ \ $MultiMAE $\ \ $& $\ \ $MAE$\ \ $ & $\ \ $DINO$\ \ $ & $\ \ $MoCoV3$\ \ $ \\
\hline
Test AUC: $\ \ $ & 88.52 & 85.53 & 85.63 & 84.98\\
\end{tabular}
\end{table}

\subsubsection{Effect of Image Resolution}
We now evaluate the impact of different image resolutions. The results are shown in Table \ref{tab:resolution_ablation}. We used 224x224 to achieve strong performance with relatively low latency, which is important for industrial systems. Resolution and the number of modalities impact the number of tokens, and, thus, affect latency. Further hyperparameter tuning may improve performance for higher resolutions. 

\begin{table}[h]
\centering
\begin{tabular}{c|c|c|c|c|c}
Resolution & 96×96 & 144×144 & 224×224 & 304×304 & 400×400 \\
\hline
AUC & 89.43 & 90.18 & 90.60 & 90.54 & 90.13 \\
Time (ms) & 152 ms & 222 ms & 687 ms & 1,780 ms & 4,320 ms \\
\end{tabular}
\caption{Model performance and inference time for different resolutions (batch size = 128)}
\label{tab:resolution_ablation}
\end{table}

\subsubsection{VAE Representations}
we also performed experiments testing a BEiT \cite{bao2021beit} model using VAEs as tokenizers. We downloaded a pretrained checkpoint and finetuned using the same setting. BEiT given RGB only results in 85.61 test AUC, and 87.70 given stacked channels. This is slightly better than MAE pretrained on generic data with stacked channels (85.98), but weaker than MultiMAE pretrained with in-domain data and separate visual modalities (90.60). This supports the paper's finding  that the MAE objective is not the most important factor. It is the in-domain pretraining, encoder finetuning and the multi-modality that improve performance. MultiMAE can easily be replaced with other multi-modal encoders and this may eventually result in better performance. 

\subsubsection{Pretrain on Items, Finetune on Packages}
We also tried pretrain on item picking data and then run pick success training on package data. This creates a significant distribution shift. We achieve 87.40 test AUC when further finetuning the encoder on package picking data but 85.43 when freezing the encoder. This is a lower score than pretraining on package picking data (88.28) but higher than without any pretraining (84.54). This shows that: (1) the representation learned from multimodal pretraining on a different data distribution can still be useful; and (2) further supports that in-domain pretraining-finetuning improves performance for challenging industrial settings.

\subsubsection{Latency}
On a single NVIDIA A10G GPU with 12 AMD EPYC 7R32 CPU cores for dataloader workers, consider a batch of 128 test data examples (pick candidates) to evaluate, it takes about 197 ms if the input has one visual modality, 409 ms for two, and 687 ms for three modalities. Note that the latency will be lower if there are a smaller number of pick candidates. Using different resolutions also affect latency. We plan to explore optimizing this further in future work.

\end{document}